\documentclass[11pt]{article}

\usepackage[final]{acl}

\usepackage{times}
\usepackage{latexsym}

\usepackage[T1]{fontenc}

\usepackage[utf8]{inputenc}

\usepackage{microtype}

\usepackage{inconsolata}

\usepackage{graphicx}

\usepackage{multirow}
\usepackage{booktabs} 
\usepackage{array}    
\usepackage{caption}  
\usepackage{amsmath}
\usepackage{amssymb}
\usepackage{tabularx}
\usepackage[table]{xcolor}

\title{MemPO: Self-Memory Policy Optimization for Long-Horizon Agents}

\author{
 \textbf{Ruoran Li\textsuperscript{1}},
 \textbf{Xinghua Zhang\textsuperscript{2}},
 \textbf{Haiyang Yu\textsuperscript{2}},
 \textbf{Shitong Duan\textsuperscript{2}},
 \textbf{Xiang Li\textsuperscript{2}},
\\
 \textbf{Wenxin Xiang\textsuperscript{1}},
 \textbf{Chonghua Liao\textsuperscript{1}},
 \textbf{Xudong Guo\textsuperscript{2}},
 \textbf{Yongbin Li\textsuperscript{2*}},
 \textbf{Jinli Suo\textsuperscript{1*}},
\\
 \textsuperscript{1}Tsinghua University
\\
 \textsuperscript{2}Tongyi Lab, Alibaba Group,
\\
 \small{
    \href{mailto:email@domain}{lrr24@mails.tsinghua.edu.cn,
   zhangxinghua.zxh@alibaba-inc.com, 
   }
   }
   \\
 \small{
   \href{mailto:email@domain}
   {
   shuide.lyb@alibaba-inc.com,
   jlsuo@tsinghua.edu.cn
   }
   }
}

\begin{document}
\maketitle
\begin{abstract}
Long-horizon agents face the challenge of growing context size during interaction with environment, which degrades the performance and stability. Existing methods typically introduce the external memory module and look up the relevant information from the stored memory, which prevents the model itself from proactively managing its memory content and aligning with the agent's overarching task objectives. To address these limitations, we propose the self-memory policy optimization algorithm (\textbf{MemPO}), which enables the agent (policy model) to autonomously summarize and manage their memory during interaction with environment. By improving the credit assignment mechanism based on memory effectiveness, the policy model can selectively retain crucial information, significantly reducing token consumption while preserving task performance. Extensive experiments and analyses confirm that MemPO achieves absolute F1 score gains of 25.98 over the base model and 7.1 over the previous SOTA baseline, while reducing token usage by 67.58\% and 73.12\%. The code is released at \url{https://github.com/TheNewBeeKing/MemPO}.
\end{abstract}

\section{Introduction}
As large language models (LLMs) continue to evolve, LLM agents are becoming increasingly proficient in addressing more complex problems. In areas such as deep research \citep{zhang_deep_2025,zheng_deepresearcher_2025}, data analysis \citep{hong_data_2025}, and vibe coding \citep{zhang_codeagent_2024,islam_mapcoder_2024,ho_verilogcoder_2025}, they have showcased remarkable performance. 
Long-horizon decision-making has always been one of the core capabilities for agents to solve complex user queries.

Currently, the dominant method for the agent-environment interaction is ReAct paradigm \citep{yao_react_2022}. The feedback from the environment is attached to the previous interaction history and is used as a prompt, which then determines the next course of action. However, this approach causes the context to grow linearly with each round of interaction, resulting in longer contexts when tackling more complex problems, and presenting several challenges. Firstly, current LLMs have relatively limited context window sizes, which impose an explicit upper bound on the number of interactions. 
Secondly, long contexts lead to excessively high token costs, which impedes the widespread adoption of agent systems in practical scenarios. Furthermore, excessively long contexts can lead to the ``lost in the middle'' phenomenon \citep{liu_lost_2023}, which degrades the model's ability, thereby reducing the overall performance of the agent.

To address this challenge, a growing body of research is focusing on agent memory, with the aim of providing LLMs with historical interaction records to reduce the need for the entire context. Currently, the mainstream solution involves designing a memory module as an external knowledge database to maintain the agent's interaction history. When the memory module is accessed, relevant historical information is retrieved and integrated into the prompt \citep{borgeaud_improving_2022,gao_retrieval-augmented_2024,lewis_retrieval-augmented_2020} based on the retrieval-augmented generation technique (RAG). 
However, the offline memory context compression method lacks the capacity for joint optimization oriented toward the agent task execution, making it difficult to effectively align with the agent's overarching task objectives.
As a result, the model's memory retrieval remains passive, rather than leveraging its own capabilities to proactively select and organize information, and the latter would facilitate more effective task completion. 

\begin{figure}[tp]
    \centering    
    \includegraphics[width=0.48\textwidth]{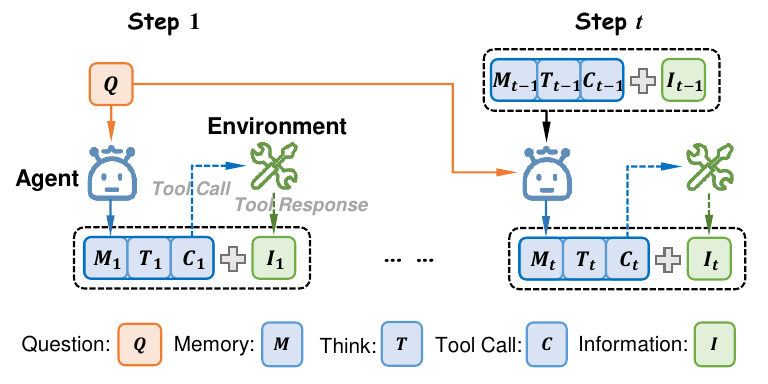}
    \caption{The self-memory inference process of our method, which only uses the previous step interaction for next step input with \texttt{<mem>} action.}
    \label{fig_intro-method}
    \vspace{-2mm}
\end{figure}

To this end, we formalize the agent interaction paradigm as autonomously refining and organizing historical information, while simultaneously reasoning and invoking tools with three actions \verb|<mem>|, \verb|<think>|, and \verb|<tool_call>|. In this paradigm, the agent itself proactively compresses and reorganizes long‑horizon historical information for the next step of interaction, making memory management an intrinsic part of its capabilities, as shown in Figure~\ref{fig_intro-method}.
To further enhance this ability, we propose self-memory policy optimization (\textbf{MemPO}), which incorporates the trajectory-level and memory-level information into advantage estimation to optimize the \verb|<mem>| action for agent with task-objective awareness.
Concretely, the tokens output by the agent are assigned trajectory-level advantages, and in each step of the interaction, the tokens of the \verb|<mem>| action additionally take into account memory-level advantages, effectively alleviating the credit assignment problem in long-horizon, multi-turn interactions. In terms of dense rewards for \verb|<mem>| action in each step, the conditional probability of the answer given \verb|<mem>| content is designed to measure the quality of the \verb|<mem>| action.
Our contributions are as follows:

\begin{itemize}
    \item We render memory management an intrinsic part of the agent's own capabilities that differs from external memory modules, achieving joint optimization of long-horizon memory, reasoning, and tool invocation.
    \item We propose \textbf{MemPO}, a self-memory policy optimization algorithm, which effectively addresses credit assignment and steers the \verb|<mem>| action toward retaining the most relevant information for solving the task.
    \item Extensive experiments on five long-horizon benchmarks confirm the efficacy of MemPO with 25.98\% and 7.1\% absolute F1 gains over the base model and previous SOTA, 67.58\% and 73.12\% reductions in token usage.
\end{itemize}

\section{Related Works}
\subsection{Memory for LLM agents}
In recent years, researchers have introduced external memory and experience systems to address the limitations of LLM context windows \citep{xuAMEMAgenticMemory2025a,chhikaraMem0BuildingProductionReady2025a,zhengSynapseTrajectoryasExemplarPrompting2024,packerMemGPTLLMsOperating2024,zhongMemoryBankEnhancingLarge2023,zhang2026expseek, liao2025exploring}. MemGPT \citep{packerMemGPTLLMsOperating2024} proposes an operating-system-inspired virtual memory management framework that employs multiple memory hierarchies to manage contextual information. Mem0 \citep{chhikaraMem0BuildingProductionReady2025a} enhances memory capacity through dynamic extraction, consolidation, and retrieval of conversational information. Despite their effectiveness in specific domains, most of these approaches rely on fixed workflows and limited optimization flexibility. They typically fail to support flexible cross-stage joint optimization, which constrains the adaptability and scalability of the overall system.

\subsection{RAG in Memory System}
RAG has emerged as a powerful approach for enhancing LLM by incorporating external knowledge sources to improve model performance \citep{borgeaud_improving_2022,gao_retrieval-augmented_2024,lewis_retrieval-augmented_2020}. In existing memory systems, the retrieval of relevant memory fragments is predominantly implemented based on RAG. While this approach can efficiently surface relevant information in certain scenarios, its major limitation lies in the lack of flexibility and end-to-end joint optimization. Specifically,  retrieval relies solely on embedding similarity between the query and chunks, which does not necessarily yield information that is most useful for solving the target problem.

\subsection{RL for LLM Agents}
The recent success of reinforcement learning  methods in LLMs has established RL as a central tool to enhance LLM-based agents to solve increasingly complex tasks \citep{jinSearchR1TrainingLLMs2025,chenReSearchLearningReason2025,zheng_deepresearcher_2025,kang2025entropy,nie2026attnpo,liao2025moa}. However, relatively few studies have explored applying RL to the optimization of agent memory. Existing approaches exhibit notable limitations. For example, 
MEM1 \citep{zhouMEM1LearningSynergize2025} integrates memory into the reasoning process and applies RL optimization for the policy model. However, it does not explicitly design objectives for memory optimization, which can lead to suboptimal memory representations. In contrast, our method introduces a dedicated credit assignment mechanism for memory rewards, encouraging the model to retain information that is most relevant for solving the target task.

\begin{figure*}
    \centering    
    \includegraphics[width=1\textwidth]{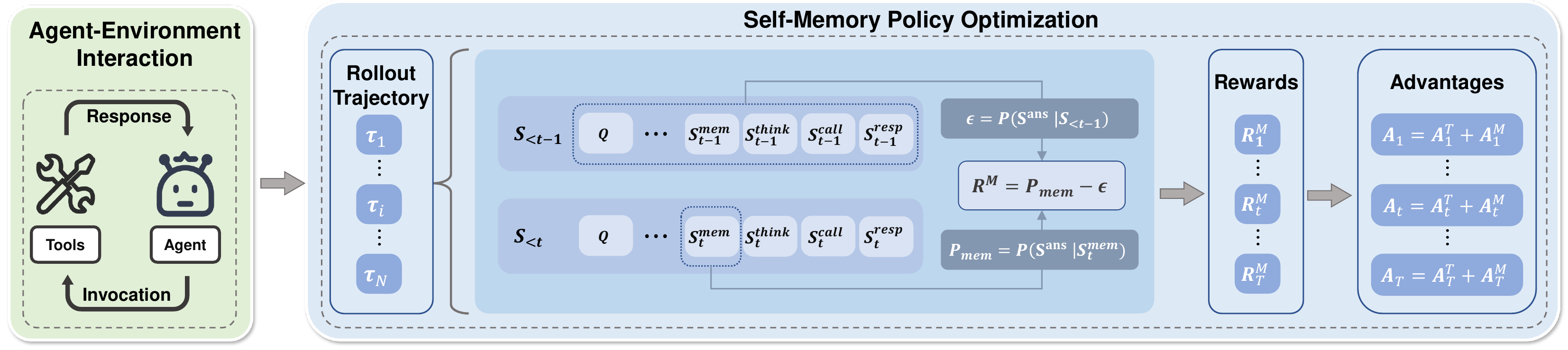}
    \caption{Overview of MemPO. At step $t$ of any trajectory $\tau_i$, the context is represented as $\{s_t^{mem}, s_t^{think}, s_t^{call}, s_t^{resp}\}$. The memory reward $R^M$ is calculated using conditional probabilities and contributes to the advantage $A^M$. The final advantage is the sum of $A^M$ and the trajectory-level advantage $A^T$. During inference, only the previous step’s content is used as context, discarding earlier information.
}
    \label{fig_main-method}
\end{figure*}

\section{Preliminaries}

\subsection{Task Formulation}
\label{sec3-1:task-formulation}
Given a question-answer pair $(q, a_{gt})$, when an LLM-based agent is tasked with solving the question $q$, it interacts with the external environment through multiple rounds of reasoning and tool invocation to acquire the information required for problem solving. If the agent completes the task after $T$ steps, a full trajectory can be denoted as $\tau = \{s_1, s_2, \ldots, s_T\}$. 

Each state $s_t$ is further decomposed into $\{s_t^{mem},$ $ s_t^{think}, s_t^{call}, s_t^{resp}\}$. Specifically, $s_t^{mem}$ represents the model-generated summary of effective information from previous outputs $s_{<t}$, which is enclosed by $\texttt{<mem> </mem>}$. $s_t^{think}$ corresponds to the model's reasoning process and is wrapped by $\texttt{<think> </think>}$. $s_t^{call}$ denotes the invocation of external tools by the model, which is represented as $\texttt{<tool\_call> </tool\_call>}$. $s_t^{resp}$ captures the information returned by the tool and is enclosed by $\texttt{<information> </information>}$. Once the agent has gathered sufficient information to answer the question $q$, it produces a predicted answer $a_{pred}$, wrapped by $\texttt{<answer>  </answer>}$. 

\subsection{Group Relative RL}

In reinforcement learning for LLM, a class of
group-based methods, exemplified by Group Relative Policy Optimization (GRPO) \citep{shaoDeepSeekMathPushingLimits2024}, abandon
per-trajectory value function modeling and instead performs relative
comparison within a batch of candidate trajectories. Concretely, for a
given task input $q$, the policy $\pi_{\theta_{\text{old}}}$ generates
$N$ complete trajectories
$\{\tau_1,\tau_2,\dots,\tau_N\}$ in one shot, and each trajectory is
assigned a scalar return $R(\tau_i)$ that measures the overall quality
of the generated outcome. The algorithm then relies solely on
statistics within this group to construct advantages, without
explicitly learning a value network:
\begin{equation}
  A(\tau_i)
  = \operatorname{GroupAgg}\Bigl(\{R(\tau_j)\}_{j=1}^N,\, i\Bigr),
\end{equation}
where $\operatorname{GroupAgg}(\cdot)$ is the aggregation operator based
on normalization, or pairwise comparison. 

The design bypasses the instabilities of value function estimation and reduces it to modeling relative preferences among a set of candidate answers. In large-scale LLM training,
group-based methods can reduce the memory overhead of extra networks, making them an efficient alternative for RL training.

\subsection{Behavior Cloning}
To enable the model to better follow the action format, we first adopt GPT-4.1 \citep{openai_gpt-4_2024} to perform inference on the publicly available training dataset from the work of \citep{tang_beyond_2025}, filter out trajectory with incorrect answers, and finally generate approximately 10K trajectories following the predefined action format in \S~\ref{sec3-1:task-formulation}. Based on these trajectory data, we fine-tune the LLM and provide a promising starting point for self-memory policy optimization.

\section{Self-Memory Policy Optimization}
As mentioned above, memory is introduced to address long contexts of agents by removing irrelevant information and retaining key details. Vanilla GRPO computes rewards based on answer correctness and uses trajectory-level advantages, where tokens within the same trajectory share the same reward. It provides the sparse rewards and limited guidance for memory generation, as the correctness of the final answer can not directly reflect the quality of each $\texttt{<mem>}$ action during the interaction.

To address this, we propose \textbf{MemPO}, a self-memory policy optimization algorithm. We design a novel advantage computation method that, in addition to trajectory-level advantages, evaluates the information content of memory within $\texttt{<mem> </mem>}$ at each step and computes an additional advantage, ensuring memory remains concise while preserving important information.

\subsection{Advantages of Global Trajectory}
\label{adv-traj}
We first evaluate the trajectory format and the accuracy of the final answer to provide a coarse-grained assessment of the overall trajectory quality. Suppose that for a single training sample, we perform $N$ rollouts and assign an overall score to each resulting trajectory, denoted as a group:
\begin{equation}
\begin{aligned}
G^T &= \{(\tau_1, R^T(\tau_1)), (\tau_2, R^T(\tau_2)), \\
& \ldots, (\tau_N, R^T(\tau_N))\}.
\end{aligned}
\end{equation}
where $\tau_i$ denotes a trajectory, and $R^T(\tau_i)$ represents the trajectory-level reward. In our method, the reward consists of evaluations of both the output format and the correctness of predicted answer. 
Concretely, the reward is set to 1 if and only if the predicted answer is correct and output format is proper; otherwise, it is set to 0.

To assess the global relative quality of each trajectory within the group, we adopt the advantage calculation strategy from GRPO \citep{shaoDeepSeekMathPushingLimits2024}, which normalizes the total reward using the mean and standard deviation computed over the group:
\begin{equation}
A^{T}(\tau_i) =
\frac{R^T(\tau_i) - \operatorname{mean}\left(\{R^T(\tau_j)\}_{j=1}^N\right)}
{\operatorname{std}\left(\{R^T(\tau_j)\}_{j=1}^N\right)}.
\end{equation}


\subsection{Advantages of Informative Memory}
\label{adv-info-mem}



According to the probabilistic formulation of LLMs, the output of model is characterized as conditional probabilities given the preceding context, i.e., $\pi_{\theta}(s_t \mid q, s_{<t})$ \citep{vaswani_attention_2017}. As demonstrated and exploited in previous work \citep{wang_information_2025,kim_re-rag_2024,lewis_retrieval-augmented_2020}, if the context $s_{<t}$ contains sufficient information to solve the problem, the probability that $s_t$ is sampled as the answer-generation step $s^{\rm ans}$ will be relatively high. 
Similarly, this insight suggests that for an arbitrary context $s^{\rm any}$, a higher value of $\pi_{\theta}(s^{\rm ans} \mid q, s^{\rm any})$ indicates that $s^{\rm any}$ contains more key information relevant to solving the question $q$, thereby increasing the model's confidence in generating the correct answer. Consequently, conditional probability can be used as a quantitative measure of the effective information content contained in a given context.

Based on this intuition, we design a step-level reward for the memory ($\texttt{<mem>}$ action) generated at each interaction step, which reflects the quality of effective information retained in memory:
\begin{equation}
\begin{aligned}
 R^M\bigl(\tau_i(s^{mem}_t)\bigr) = &P\!\left[s^{\rm ans} \mid \tau_i(s^{mem}_t)\right] - \epsilon,\\
& 1 \leq i \leq N,\; 1 \leq t \leq T.
\end{aligned}
\label{eq_mem-reward}
\end{equation}
where $\tau_i(s^{mem}_t)$ denotes the memory content within $\texttt{<mem>}$ action in step $t$ of trajectory $\tau_i$, and $s^{\rm ans} = \{a_1, a_2, \dots, a_L\}$ represents the correct answer string, where $a_l$ denotes the $l$-th token of the answer. The $\epsilon$ represents $P\!\left[s^{\rm ans} \mid \tau_i(s_{<t})\right]$, which serves as a bias term. The term $\tau_i(s_{<t})$ corresponds to the trajectory of the first $t$-1 steps of $\tau_i$. The operator $P(\cdot)$ denotes a posterior-probability-based measure. 
Specifically, $P\!\left[s^{\rm ans} \mid \tau_i(s^{mem}_{t})\right]$ can be represented as:
\begin{equation}
\sqrt[L]{\prod_{l=1}^{L}
\pi_{\theta}\!\left(a_l \mid q, \tau_i(s^{mem}_{t}), a_{<l}\right)}.
\end{equation}
where $\pi_{\theta}(a_l \mid q, \tau_i(s^{mem}_{t}), a_{<l})$ denotes the probability of generating token $a_l$ given the user query, the trajectory prefix up to step $t-1$, and the preceding answer tokens $a_{<l}$.

Under this formulation, a larger value of $R^M\!\left(\tau_i(s^{mem}_t)\right)$ indicates that the memory generated at step $t$ provides a more effective summary of the trajectory up to the first $t$-1 steps, and better preserves contextual information that is relevant to generating the correct answer.

Based on the reward formulation above, the resulting memory group can be expressed as:
\begin{equation}
\begin{aligned}
G^M = \bigl\{ &\left(\tau_i(s^{mem}_t), R^M(\tau_i(s^{mem}_t))\right) \\
& | 1 \leq i \leq N,\; 1 \leq t \leq T \bigr\}.
\end{aligned}
\end{equation}
We then normalize the rewards using the group-wise mean and standard deviation to obtain the corresponding advantages $A^{M}\!\left(\tau_i(s^{mem}_t)\right)$:
\begin{equation}
A^{M}\!\left(\tau_i(s^{mem}_t)\right) = \frac{
R^M\!\left(\tau_i(s^{mem}_t)\right)
-
\operatorname{M}\!\left(\tau_i(s^{mem}_t)\right)
}{
\operatorname{std}\left(\{R^M(\tau_i(s^{mem}_t))\}\right)
}.
\end{equation}
where $\operatorname{M}(\tau_i(s^{mem}_t))$ denotes the mean reward within the same group, defined as:
\begin{equation}
\begin{aligned}
 \operatorname{M}(&\tau_i(s^{mem}_t)) = \operatorname{mean}\bigl(\bigl\{R^M\!\left(\tau_i(s^{mem}_t)\right) \\
&|\;\bigl(\tau_i(s^{mem}_t),  R^M(\tau_i(s^{mem}_t))\bigr) \in G^M\bigr\}\bigr).
\end{aligned}
\end{equation}

The advantage $A^{M}\!\left(\tau_i(s^{mem}_t)\right)$ provides a quantitative assessment of memory quality, enabling finer-grained supervision over the model-generated memory content.

\subsection{Combination of Advantages}
The final token-level advantage is obtained by combining the two types of advantages in \S~\ref{adv-traj} and \S~\ref{adv-info-mem}. Let the $k$-th token of the $i$-th trajectory $\tau_i$ in a group be denoted as $\tau_{i,k}$. The advantage assigned to this token $A_{i,k}$ is defined as:
\begin{equation}
A_{i,k}=\begin{cases}
A^T(\tau_i) \!\!+ \!\!A^M\!\left(\tau_i(s^{mem}_t)\right),\!\!\!\!
& \tau_{i,k} \!\in\! \tau_i(s^{mem}_t) \\
A^T(\tau_i),
& \text{otherwise}.
\end{cases}
\end{equation}

That is, when $\tau_{i,k}$ corresponds to a token within the memory segment ($\texttt{<mem>}$ action), its advantage is given by the sum of the trajectory-level advantage and the memory-level advantage; otherwise, only the trajectory-level advantage $A^T$ is used. In this way, tokens belonging to memory receive richer and more explicit feedback signals, which more effectively guide the rollout process for memory generation.

\subsection{Policy Optimization and Inference}
\textbf{Optimization.}
The policy optimization objective is to maximize $\mathcal{J}(\theta)$, written as:
\begin{equation}
\begin{aligned}
 \mathcal{J}(\theta) &= \mathbb{E} \Bigg[\frac{1}{N}\sum_{i=1}^{N}\frac{1}{|\tau_i|}\sum_{k=1}^{|\tau_i|}\min\Big(\gamma_{i,k} A_{i,k}, \operatorname{clip}(\gamma_{i,k}\\
& ,1 - \epsilon,1 + \epsilon)A_{i,k}\Big)- \beta\,\mathcal{D}_{\mathrm{KL}}\!\left(\pi_\theta \,\|\, \pi_{\mathrm{ref}}\right)\Bigg].
\end{aligned}
\end{equation}
where $\gamma_{i,k}$ is the importance sampling ratio:
\begin{equation}
\frac{\pi_\theta(\tau_{i,k} \mid q, \tau_{i,<k})}
{\pi_{\theta_{\mathrm{old}}}(\tau_{i,k} \mid q, \tau_{i,<k})}.
\end{equation}
where ${q \sim p(Q),\, \{\tau_i\}_{i=1}^{N} \sim \pi_{\theta_{\mathrm{old}}}}$. $p(Q)$ denotes the distribution of queries in the training set, and $\beta$ controls the weight of the KL-divergence regularization term.

\textbf{Inference.} In vanilla ReAct framework, the $t$-th step inference is denoted as $\pi_{\theta}(s_t \mid q, s_{<t})$. In our method, since $s_{t-1}^{mem}$ contains the effective information of $s_{<t-2}$, we replace $s_{<t}$ with $s_{t-1}^{mem}$ as the inference context, represented as $\pi_{\theta}(s_t \mid q, s_{t-1}^{mem})$.

\begin{table*}[htbp]
  \centering
  \caption{The accuracy and token consumption for multi-objective tasks of baselines.\textbf{Text with bold} means SOTA.}
  \resizebox{\textwidth}{!}{%
    \begin{tabular}{l|cccccccccc|cccc}
    \toprule
    \toprule
    \rowcolor[rgb]{ .949,  .949,  .949} \multicolumn{15}{c}{\textbf{Local Wiki Search}} \\
    \midrule
    \midrule
    \multirow{2}[2]{*}{Model} & \multicolumn{2}{c}{\textbf{2-objective}} & \multicolumn{2}{c}{\textbf{4-objective}} & \multicolumn{2}{c}{\textbf{6-objective}} & \multicolumn{2}{c}{\textbf{8-objective}} & \multicolumn{2}{c|}{\textbf{10-objective}} & \multicolumn{4}{c}{\textbf{Avg}} \\
          & F1    & EM    & F1    & EM    & F1    & EM    & F1    & EM    & F1    & EM    & F1 $\uparrow$    & EM $\uparrow$   & TT $\downarrow$   & PT $\downarrow$ \\
    \midrule
    Qwen2.5 (ReAct) & 33.73  & 25.60  & 10.59  & 7.00  & 5.37  & 4.00  & 5.92  & 4.25  & 2.63  & 1.96  & 11.65  & 8.56  & 3.64  & 0.61  \\
    ReSearch & 47.40  & 36.00  & 24.13  & 16.70  & 20.84  & 15.93  & 10.80  & 7.70  & 5.08  & 3.56  & 21.65  & 15.98  & 3.29  & 0.71  \\
    DeepResearcher & 30.94  & 24.70  & 24.52  & 18.10  & 13.88  & 10.73  & 9.12  & 6.85  & 5.07  & 3.52  & 16.71  & 12.78  & 4.29  & 0.77  \\
    A-MEM & 33.24  & 25.20  & 13.71  & 10.10  & 9.80  & 7.07  & 6.86  & 5.10  & 5.56  & 3.60  & 13.83  & 10.21  & 2.62  & 0.38  \\
    MEM1  & 47.74  & 37.10  & 26.51  & 18.90  & 18.81  & 14.07  & 19.04  & 13.55  & 19.61  & 13.36  & 26.34  & 19.40  & 1.38  & 0.20  \\
    GRPO (w/o $\texttt{<mem>}$) & 54.57  & 42.95  & 38.31  & 28.60  & 29.78  & 22.60  & 18.97  & 13.65  & 11.01  & 7.84  & 30.53  & 23.13  & 4.39  & 0.81  \\
    \textbf{MemPO (Ours)}  & \textbf{56.47 } & \textbf{46.15 } & \textbf{42.75 } & \textbf{31.90 } & \textbf{34.32 } & \textbf{26.93 } & \textbf{30.48 } & \textbf{23.70 } & \textbf{24.15 } & \textbf{18.16 } & \textbf{37.63 } & \textbf{29.37 } & \textbf{1.18 } & \textbf{0.18 } \\
    \midrule
    \midrule
    \rowcolor[rgb]{ .949,  .949,  .949} \multicolumn{15}{c}{\textbf{Online Web Search}} \\
    \midrule
    \midrule
    Qwen2.5 (ReAct) & 45.71  & 33.60  & 16.83  & 11.80  & 12.12  & 9.07  & 9.66  & 6.80  & 7.25  & 4.68  & 18.31  & 13.19  & 3.14  & 0.34  \\
    ReSearch & 51.17  & 39.20  & 29.92  & 21.10  & 25.21  & 18.73  & 17.09  & 12.50  & 10.37  & 7.44  & 26.75  & 19.79  & 2.17  & 0.40  \\
    MEM1  & 50.56  & 39.60  & 30.43  & 22.00  & 21.67  & 16.20  & 19.48  & 14.30  & 18.06  & 12.12  & 28.04  & 20.84  & 0.96  & 0.14  \\
    \textbf{MemPO (Ours)}  & \textbf{57.40 } & \textbf{45.20 } & \textbf{41.42 } & \textbf{30.20 } & \textbf{37.92 } & \textbf{28.60 } & \textbf{34.30 } & \textbf{25.80 } & \textbf{22.92 } & \textbf{16.32 } & \textbf{38.79 } & \textbf{29.22 } & \textbf{0.86 } & \textbf{0.12 } \\
    \bottomrule
    \bottomrule
    \end{tabular}%
    }
  \label{tab_multi-obj_acc-token-num}%
\end{table*}%

\section{Experiments}

\subsection{Benchmarks}
To evaluate the effectiveness of our approach, following the method of MEM1 \citep{zhouMEM1LearningSynergize2025}, we test on multi-objective tasks, where the number of interaction rounds required for the agent to solve a problem is significantly higher compared to single-objective tasks. This allows us to better assess the performance of our method in scenarios with long contexts. Additionally, we can observe the changes in agent performance by progressively increasing the number of objectives. We created a 2-objective task test set by combining queries from the validation sets of the HotpotQA \citep{yang_hotpotqa_2018} and NQ \citep{kwiatkowski_natural_2019} QA datasets, and synthesized test sets with more objectives using the HotpotQA validation set. We conducted tests under both local wiki search engine and web search engine scenarios to enhance the credibility of the experiments.

Following previous work \citep{zhouMEM1LearningSynergize2025}, we use the \textbf{F1} score as a criterion for word-level matching, and Exact Match (\textbf{EM}) for exact matching. Furthermore, to evaluate the token consumption of the agent when solving a problem, we use the total number of tokens consumed to solve a question (\textbf{TT}), as well as the maximum number of tokens (peak tokens) consumed in a single step (\textbf{PT}).

\subsection{Baselines}
We compare our method with various baselines. For prompt-based baselines, we use ReAct \citep{yao_react_2022}. For agentic RL-based baselines, we adopt DeepResearcher \citep{zheng_deepresearcher_2025} and ReSearch \citep{chenReSearchLearningReason2025}. For agent memory-related baselines, we use RL-based method MEM1 \citep{zhouMEM1LearningSynergize2025} and RAG-based method A-MEM \citep{xuAMEMAgenticMemory2025a}. Additionally, we also trained a model without memory using GRPO in the exact same environment as a baseline. To ensure fairness, all methods use the 7B model from the Qwen2.5 series as the base model.

\subsection{Implementation Details}
We first performed inference using GPT-4.1 \citep{openai_gpt-4_2024} on the dataset from the work of \cite{tang_beyond_2025}, and obtained approximately 10k trajectories containing memory. We then fine-tuned the model for one epoch using these data to enhance its ability to follow instructions related to the memory component. Simultaneously, we removed the memory component from the trajectories to use them for fine-tuning the baseline model, which was trained using GRPO and does not include memory, ensuring fairness in the comparison.

In the RL phase, we followed MEM1 and used the 2-objective task synthesized from HotpotQA and NQ as part of the training set. And we randomly sampled a subset from both datasets as another part of training set. The rollout group size $N$ for group-based RL methods is set to 16, with a batch size of 128 and a learning rate of 1e-6. The maximum number of interaction rounds is set to 16. During training, we use the local wiki search engine  as the search tool.

\begin{figure*}
    \centering
    \begin{minipage}{0.48\textwidth}
        \centering
        \includegraphics[width=\textwidth]{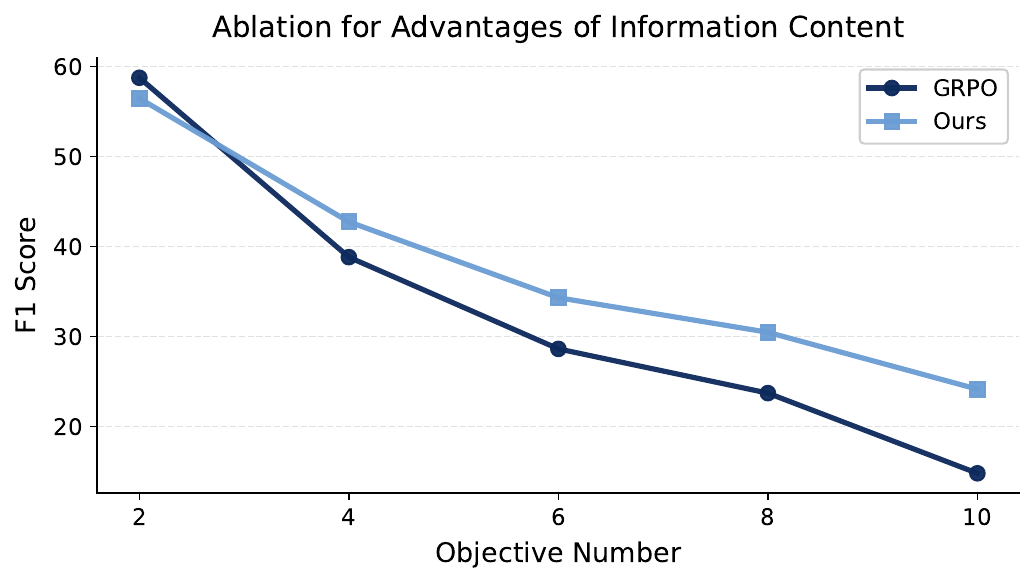}
    \end{minipage}
    \begin{minipage}{0.48\textwidth}
        \centering
        \includegraphics[width=\textwidth]{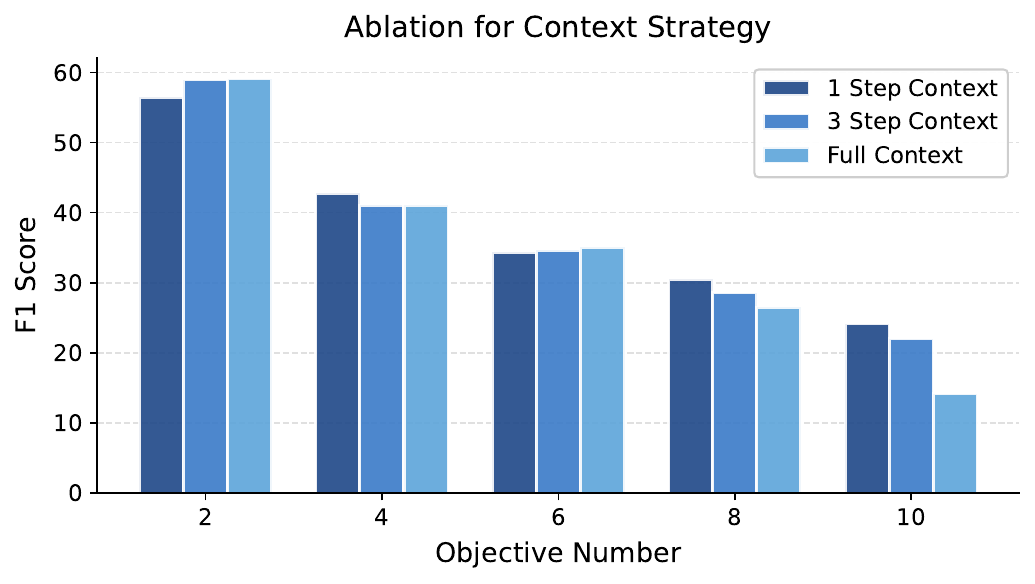}
    \end{minipage}
    \caption{The results of ablation study.}
    \label{fig_ablation}
\end{figure*}

\begin{figure*}
    \centering    
    \includegraphics[width=1\textwidth]{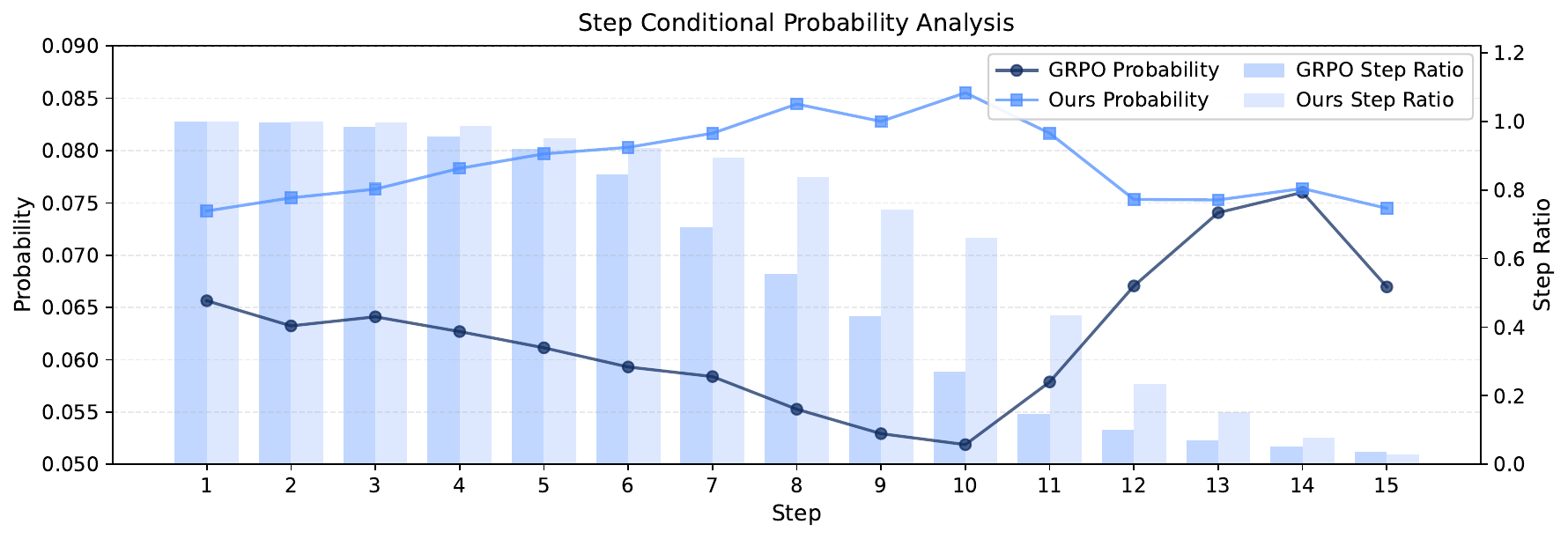}
    \caption{The result of step conditional probability analysis. }
    \label{fig_step-prob}
\end{figure*}

\subsection{Experimental Results}

\textbf{Multi-objective task.} The results of each baseline on the multi-objective task are shown in Table~\ref{tab_multi-obj_acc-token-num}. We selected tasks with 4, 6, 8, and 10 objectives as progressively harder task groups and recorded F1 and EM for the answers from each baseline as precision metrics. Among the baselines, MEM1, A-MEM, and our method use truncated contexts, meaning the model only has access to the previous step of interactions, while the other baselines use the complete context. Additionally, we also recorded the total number of tokens required to solve a single problem (TT) and the peak token consumption per step (PT) during the model's execution. The more detailed results are presented in Appendix Table~\ref{tab_multi-obj_token-num}.

\textbf{Conditional probability analysis.} 
To investigate the impact of our reward design on the model, we performed a statistical analysis of the true values of $P[s^{\rm ans} \mid s^{mem}]$ during inference on the 10-objective task. We compared the results of models trained with vanilla GRPO and our method. Specifically, let the dataset size be $M$, and denote the memory component at step $t$ of the $m$-th trajectory as $\tau_m(s^{mem}_t)$, with the corresponding ground truth answer string being $\tau_m(s^{\rm ans})$. Figure~\ref{fig_grouped-prob} shows the grouped results of $P[\tau_m(s^{\rm ans}) \mid \tau_m(s^{mem}_t)]$, where the x-axis represents the group values of the conditional probability, and the y-axis shows the proportion of memory samples in that range relative to the total memory samples. This distribution illustrates the conditional probability distribution of the memory produced by the model. The line graph's y-axis represents the average accuracy of trajectories whose memory falls within each group, providing insight into the relationship between accuracy and the conditional probability.

Figure~\ref{fig_step-prob} displays the aggregated results of $P[\tau_m(s^{\rm ans}) \mid \tau_m(s^{mem}_t)]$ by step. The x-axis represents the step $t$ of the memory, and the y-axis of the line graph shows the average conditional probability of the memory at step $t$ across all $M$ trajectories. This provides insight into how conditional probability evolves with the step. The histogram shows the proportion of memories at each step relative to the total number of trajectories.

\textbf{Ablation study.} To validate the effectiveness of our design, we compared the performance of the memory-enabled model trained with vanilla GRPO and our model, keeping all other conditions identical. The only difference between the two models is the inclusion of a reward specifically for memory. The results are shown in the left panel of Figure~\ref{fig_ablation}. Additionally, we evaluated our method under different context retention settings: full context, retaining 1 or 3 interaction rounds. The results of these experiments are shown in the right panel of Figure~\ref{fig_ablation}.

\subsection{Experimental Analyses} 

\textbf{Our method demonstrates remarkable performance and generalization.} As shown in Table~\ref{tab_multi-obj_acc-token-num}, we achieve SOTA performance on tasks that are much more difficult than those in the training set, and our model continues to maintain leading performance when switched to a real-world web search environment that differs from the training setup. Additionally, as presented in Table~\ref{tab_multi-obj_acc-token-num}, our model achieves SOTA performance while minimizing token consumption, thereby achieving the highest performance with the least resource usage. Furthermore, as seen in Figure~\ref{fig_ablation}, whether using full context, retaining 1 or 3 interaction rounds for inference, our method's performance remains consistently stable, showcasing strong generalization capabilities.

\textbf{The memory mechanism significantly reduces token consumption.} As shown in Appendix Table~\ref{tab_multi-obj_token-num}, token consumption for tasks solved by MEM1, A-MEM, and our method, which all incorporate the memory mechanism, is noticeably lower than that of other baselines. Taking ReSearch as an example and comparing it with our method, when the task is relatively simple, such as a 2-objective task, the token consumption is only slightly higher than ours. However, as the complexity of the task increases, the gap between the two methods becomes more pronounced. By the time the task reaches 10 objectives, the number of tokens required by our method to solve a problem is approximately 1/3 of ReSearch's, with the token peak being 1/5. This is comparable to ReSearch's token consumption on a 4-objective task. Moreover, our method not only uses truncated contexts but also provides effective guidance on the memory content, resulting in even more compact contexts compared to other memory-related methods. As a result, token consumption in our method is the lowest among all baselines.

\begin{figure*}
    \centering    
    \includegraphics[width=1\textwidth]{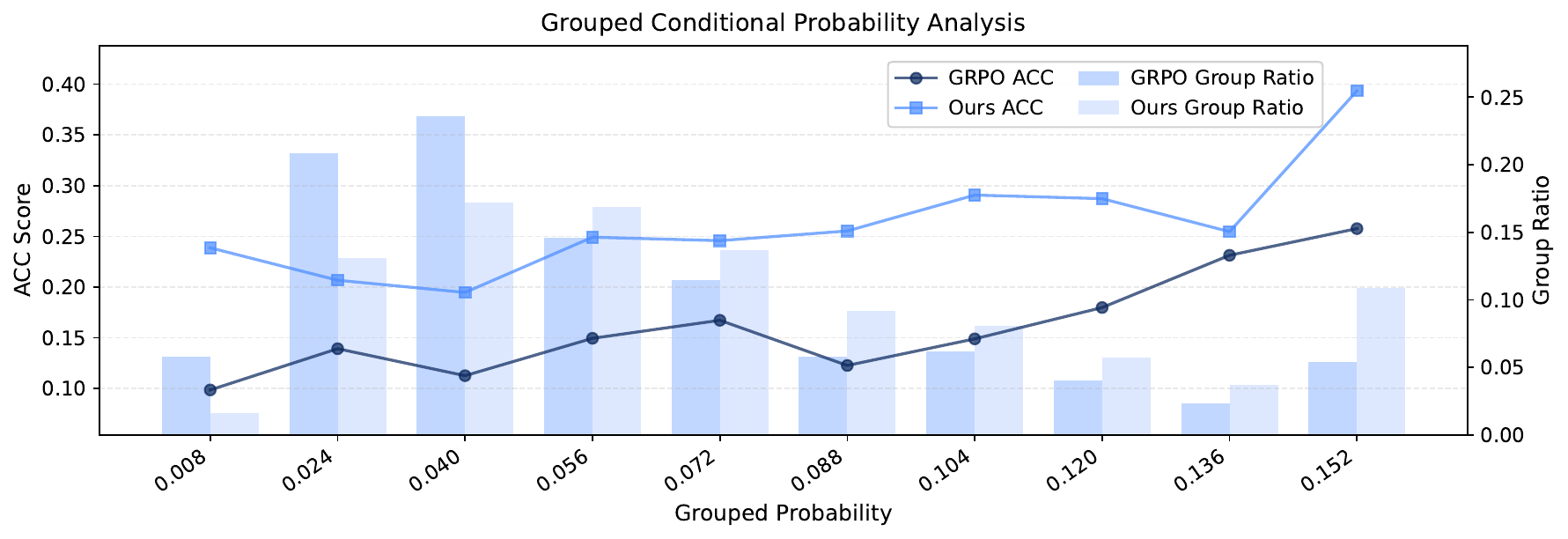}
    \caption{The result of grouped conditional probability analysis. }
    \label{fig_grouped-prob}
\end{figure*}

\textbf{The ability of memory to provide strongly relevant information is crucial for task success.} In the baseline methods, A-MEM generates memory based on a RAG approach. As mentioned in the introduction, the memory obtained by this method is not necessarily the most relevant for solving the task and contains a significant amount of redundancy. As a result, while A-MEM reduces token consumption compared to ReAct, its performance does not show a significant improvement. On the other hand, MEM1 generates memory by combining the model's summary with reasoning, creating a stronger link between memory generation and the task-solving process. This allows MEM1 to show a considerable improvement on long-horizon tasks. Furthermore, our method explicitly guides the model to retain the context that most strongly contributes to solving the problem, outperforming other memory-related baselines in all datasets.

\textbf{The number of context steps impacts performance on long-horizon tasks.} As shown in the right panel of Figure~\ref{fig_ablation}, we present the results of inference using complete context, truncated in 1 step and in 3 steps. Overall, the performance differences between these methods fluctuate within an acceptable range, with the trend showing that the more context steps used, the better the performance on short-horizon tasks, but the weaker the performance on long-horizon tasks. This effect is particularly noticeable on long-horizon tasks. We believe that this aligns with the phenomenon of attention dilution caused by long contexts, which leads to performance degradation. 


\textbf{Our reward design positively contributes to improving the effective information content.} Figure~\ref{fig_ablation} presents a performance comparison between vanilla GRPO and our method. The results show that our reward design leads to an improvement in the model's performance. We also quantitatively analyzed the information content of the memory in the trajectories of both vanilla GRPO and our method. The bar graph in Figure~\ref{fig_grouped-prob} indicates that, compared to the baseline, our method’s probability distribution is more skewed toward higher values, which contributes to greater precision of responses, as confirmed by the line graph.

Additionally, the line graph in Figure~\ref{fig_step-prob} shows that, for the first 10 steps of 10-objective task, the mean probability of our method increases as the steps progress, whereas the baseline shows a decreasing trend. We believe this reflects the more effective organization of memory by our method compared to the baseline. After 10 steps, our method’s probability starts to decrease, which is reasonable given that the typical number of search steps in a 10-objective task is around 10. If the task is not completed by this point, it suggests that some information is difficult to find and is still being searched for. In contrast, the baseline experiences more difficulty in the first 10 steps, and as seen in the bar chart, only 20\% of the search examples continue after the 10th step. We hypothesize that the few remaining examples that did not abandon exploration likely achieved relatively higher accuracy, which explains the continued increase in probability after the 10th step. 

Overall, both the final performance and the probability analysis validate that our reward design is effective and aligns with expectations.
\vspace{-2mm}
\section{Conclusion}
\vspace{-2mm}
Our method optimizes memory management for agents by introducing a novel reward design that retains only relevant information, improving task performance and reducing computational costs. By integrating memory, reasoning, and tool invocation via reinforcement learning, we achieve superior performance, especially on long-horizon tasks, with efficient token consumption. Future work will focus on optimizing memory reward design and enhancing the scalability of self-memory method in more applications.

\section*{Limitations}
Although our method shows promising performance, one potential limitation is that, due to varying tool invocation at different steps, the information content in memory naturally differs and the states are not completely equivalent across all steps in all rollout trajectories, which may introduce bias when calculating group-based advantages. While we alleviate this issue by introducing $\epsilon$ in Equation~\ref{eq_mem-reward}, more refined solutions may reduce this bias in more complex environments, and understanding the generalization to diverse real‑world settings needs further investigation.

\section*{LLM Usage Statement}
In this paper, LLMs are used to assist in polishing the grammar of the article and translating related content, without any non-compliant usage.

\section*{Acknowledgment}
This work was jointly supported by the National Key R\&D Program of China (Grant No. 2024YFF0505703) and Beijing Municipal Natural Science Foundation (Grant No. L257009) .


\bibliography{acl_latex}

@article{liao2025moa,
  title={MOA: Multi-Objective Alignment for Role-Playing Agents},
  author={Liao, Chonghua and Wang, Ke and Wu, Yuchuan and Huang, Fei and Li, Yongbin},
  journal={arXiv preprint arXiv:2512.09756},
  year={2025}
}

@article{kang2025entropy,
  title={Entropy regularizing activation: Boosting continuous control, large language models, and image classification with activation as entropy constraints},
  author={Kang, Zilin and Liao, Chonghua and Xu, Tingqiang and Xu, Huazhe},
  journal={arXiv preprint arXiv:2510.08549},
  year={2025}
}

@inproceedings{liao2025exploring,
  title={Exploring forgetting in large language model pre-training},
  author={Liao, Chonghua and Xie, Ruobing and Sun, Xingwu and Sun, Haowen and Kang, Zhanhui},
  booktitle={Proceedings of the 63rd Annual Meeting of the Association for Computational Linguistics (Volume 1: Long Papers)},
  pages={2112--2127},
  year={2025}
}

@article{zhang2026expseek,
  title={ExpSeek: Self-Triggered Experience Seeking for Web Agents},
  author={Zhang, Wenyuan and Zhang, Xinghua and Yu, Haiyang and Nie, Shuaiyi and Wu, Bingli and Yue, Juwei and Liu, Tingwen and Li, Yongbin},
  journal={arXiv preprint arXiv:2601.08605},
  year={2026}
}

@misc{nie2026attnpo,
      title={ATTNPO: Attention-Guided Process Supervision for Efficient Reasoning}, 
      author={Shuaiyi Nie and Siyu Ding and Wenyuan Zhang and Linhao Yu and Tianmeng Yang and Yao Chen and Tingwen Liu and Weichong Yin and Yu Sun and Hua Wu},
      year={2026},
      eprint={2602.09953},
      archivePrefix={arXiv},
      primaryClass={cs.CL},
      url={https://arxiv.org/abs/2602.09953}, 
}

@misc{wang_information_2025,
	title = {Information {Gain}-based {Policy} {Optimization}: {A} {Simple} and {Effective} {Approach} for {Multi}-{Turn} {LLM} {Agents}},
	shorttitle = {Information {Gain}-based {Policy} {Optimization}},
	url = {http://arxiv.org/abs/2510.14967},
	doi = {10.48550/arXiv.2510.14967},
	urldate = {2025-10-22},
	publisher = {arXiv},
	author = {Wang, Guoqing and Dai, Sunhao and Ye, Guangze and Gan, Zeyu and Yao, Wei and Deng, Yong and Wu, Xiaofeng and Ying, Zhenzhe},
	month = oct,
	year = {2025},
	note = {arXiv:2510.14967 [cs]},
}

@misc{tang_beyond_2025,
	title = {Beyond {Turn} {Limits}: {Training} {Deep} {Search} {Agents} with {Dynamic} {Context} {Window}},
	shorttitle = {Beyond {Turn} {Limits}},
	url = {http://arxiv.org/abs/2510.08276},
	doi = {10.48550/arXiv.2510.08276},
	urldate = {2025-10-22},
	publisher = {arXiv},
	author = {Tang, Qiaoyu and Xiang, Hao and Yu, Le and Yu, Bowen and Lu, Yaojie and Han, Xianpei and Sun, Le and Zhang, WenJuan and Wang, Pengbo and Liu, Shixuan and Zhang, Zhenru and Tu, Jianhong and Lin, Hongyu and Lin, Junyang},
	month = oct,
	year = {2025},
	note = {arXiv:2510.08276 [cs]},
}

@misc{mallen_when_2023,
	title = {When {Not} to {Trust} {Language} {Models}: {Investigating} {Effectiveness} of {Parametric} and {Non}-{Parametric} {Memories}},
	shorttitle = {popqa},
	url = {http://arxiv.org/abs/2212.10511},
	doi = {10.48550/arXiv.2212.10511},
	urldate = {2025-12-23},
	publisher = {arXiv},
	author = {Mallen, Alex and Asai, Akari and Zhong, Victor and Das, Rajarshi and Khashabi, Daniel and Hajishirzi, Hannaneh},
	month = jul,
	year = {2023},
	note = {arXiv:2212.10511 [cs]},
}

@article{kwiatkowski_natural_2019,
	title = {Natural {Questions}: {A} {Benchmark} for {Question} {Answering} {Research}},
	volume = {7},
	issn = {2307-387X},
	shorttitle = {Natural {Questions}},
	url = {https://direct.mit.edu/tacl/article/43518},
	doi = {10.1162/tacl_a_00276},
	language = {en},
	urldate = {2025-12-23},
	journal = {Transactions of the Association for Computational Linguistics},
	author = {Kwiatkowski, Tom and Palomaki, Jennimaria and Redfield, Olivia and Collins, Michael and Parikh, Ankur and Alberti, Chris and Epstein, Danielle and Polosukhin, Illia and Devlin, Jacob and Lee, Kenton and Toutanova, Kristina and Jones, Llion and Kelcey, Matthew and Chang, Ming-Wei and Dai, Andrew M. and Uszkoreit, Jakob and Le, Quoc and Petrov, Slav},
	month = nov,
	year = {2019},
	pages = {453--466},
}

@misc{joshi_triviaqa_2017,
	title = {{TriviaQA}: {A} {Large} {Scale} {Distantly} {Supervised} {Challenge} {Dataset} for {Reading} {Comprehension}},
	shorttitle = {{TriviaQA}},
	url = {http://arxiv.org/abs/1705.03551},
	doi = {10.48550/arXiv.1705.03551},
	urldate = {2025-12-23},
	publisher = {arXiv},
	author = {Joshi, Mandar and Choi, Eunsol and Weld, Daniel S. and Zettlemoyer, Luke},
	month = may,
	year = {2017},
	note = {arXiv:1705.03551 [cs]},
}

@inproceedings{yang_hotpotqa_2018,
	address = {Brussels, Belgium},
	title = {{HotpotQA}: {A} {Dataset} for {Diverse}, {Explainable} {Multi}-hop {Question} {Answering}},
	shorttitle = {{HotpotQA}},
	url = {https://aclanthology.org/D18-1259/},
	doi = {10.18653/v1/D18-1259},
	urldate = {2025-12-23},
	booktitle = {Proceedings of the 2018 {Conference} on {Empirical} {Methods} in {Natural} {Language} {Processing}},
	publisher = {Association for Computational Linguistics},
	author = {Yang, Zhilin and Qi, Peng and Zhang, Saizheng and Bengio, Yoshua and Cohen, William and Salakhutdinov, Ruslan and Manning, Christopher D.},
	editor = {Riloff, Ellen and Chiang, David and Hockenmaier, Julia and Tsujii, Jun'ichi},
	month = oct,
	year = {2018},
	pages = {2369--2380},
}

@misc{ho_constructing_2020,
	title = {Constructing {A} {Multi}-hop {QA} {Dataset} for {Comprehensive} {Evaluation} of {Reasoning} {Steps}},
	shorttitle = {2wiki},
	url = {http://arxiv.org/abs/2011.01060},
	doi = {10.48550/arXiv.2011.01060},
	urldate = {2025-12-23},
	publisher = {arXiv},
	author = {Ho, Xanh and Nguyen, Anh-Khoa Duong and Sugawara, Saku and Aizawa, Akiko},
	month = nov,
	year = {2020},
	note = {arXiv:2011.01060 [cs]},
}

@misc{trivedi_musique_2022,
	title = {{MuSiQue}: {Multihop} {Questions} via {Single}-hop {Question} {Composition}},
	shorttitle = {{MuSiQue}},
	url = {http://arxiv.org/abs/2108.00573},
	doi = {10.48550/arXiv.2108.00573},
	urldate = {2025-12-23},
	publisher = {arXiv},
	author = {Trivedi, Harsh and Balasubramanian, Niranjan and Khot, Tushar and Sabharwal, Ashish},
	month = may,
	year = {2022},
	note = {arXiv:2108.00573 [cs]},
}

@inproceedings{press_measuring_2023,
	address = {Singapore},
	title = {Measuring and {Narrowing} the {Compositionality} {Gap} in {Language} {Models}},
	shorttitle = {bamboogle},
	url = {https://aclanthology.org/2023.findings-emnlp.378/},
	doi = {10.18653/v1/2023.findings-emnlp.378},
	urldate = {2025-12-23},
	booktitle = {Findings of the {Association} for {Computational} {Linguistics}: {EMNLP} 2023},
	publisher = {Association for Computational Linguistics},
	author = {Press, Ofir and Zhang, Muru and Min, Sewon and Schmidt, Ludwig and Smith, Noah and Lewis, Mike},
	editor = {Bouamor, Houda and Pino, Juan and Bali, Kalika},
	month = dec,
	year = {2023},
	pages = {5687--5711},
}

@misc{openai_gpt-4_2024,
	title = {{GPT}-4 {Technical} {Report}},
	url = {http://arxiv.org/abs/2303.08774},
	doi = {10.48550/arXiv.2303.08774},
	urldate = {2026-01-05},
	publisher = {arXiv},
	author = {OpenAI and Achiam, Josh and Adler, Steven and Agarwal, Sandhini and Ahmad, Lama and Akkaya, Ilge and Aleman, Florencia Leoni and Almeida, Diogo and Altenschmidt, Janko and Altman, Sam and Anadkat, Shyamal and Avila, Red and Babuschkin, Igor and Balaji, Suchir and Balcom, Valerie and Baltescu, Paul and Bao, Haiming and Bavarian, Mohammad and Belgum, Jeff and Bello, Irwan and Berdine, Jake and Bernadett-Shapiro, Gabriel and Berner, Christopher and Bogdonoff, Lenny and Boiko, Oleg and Boyd, Madelaine and Brakman, Anna-Luisa and Brockman, Greg and Brooks, Tim and Brundage, Miles and Button, Kevin and Cai, Trevor and Campbell, Rosie and Cann, Andrew and Carey, Brittany and Carlson, Chelsea and Carmichael, Rory and Chan, Brooke and Chang, Che and Chantzis, Fotis and Chen, Derek and Chen, Sully and Chen, Ruby and Chen, Jason and Chen, Mark and Chess, Ben and Cho, Chester and Chu, Casey and Chung, Hyung Won and Cummings, Dave and Currier, Jeremiah and Dai, Yunxing and Decareaux, Cory and Degry, Thomas and Deutsch, Noah and Deville, Damien and Dhar, Arka and Dohan, David and Dowling, Steve and Dunning, Sheila and Ecoffet, Adrien and Eleti, Atty and Eloundou, Tyna and Farhi, David and Fedus, Liam and Felix, Niko and Fishman, Simón Posada and Forte, Juston and Fulford, Isabella and Gao, Leo and Georges, Elie and Gibson, Christian and Goel, Vik and Gogineni, Tarun and Goh, Gabriel and Gontijo-Lopes, Rapha and Gordon, Jonathan and Grafstein, Morgan and Gray, Scott and Greene, Ryan and Gross, Joshua and Gu, Shixiang Shane and Guo, Yufei and Hallacy, Chris and Han, Jesse and Harris, Jeff and He, Yuchen and Heaton, Mike and Heidecke, Johannes and Hesse, Chris and Hickey, Alan and Hickey, Wade and Hoeschele, Peter and Houghton, Brandon and Hsu, Kenny and Hu, Shengli and Hu, Xin and Huizinga, Joost and Jain, Shantanu and Jain, Shawn and Jang, Joanne and Jiang, Angela and Jiang, Roger and Jin, Haozhun and Jin, Denny and Jomoto, Shino and Jonn, Billie and Jun, Heewoo and Kaftan, Tomer and Kaiser, Łukasz and Kamali, Ali and Kanitscheider, Ingmar and Keskar, Nitish Shirish and Khan, Tabarak and Kilpatrick, Logan and Kim, Jong Wook and Kim, Christina and Kim, Yongjik and Kirchner, Jan Hendrik and Kiros, Jamie and Knight, Matt and Kokotajlo, Daniel and Kondraciuk, Łukasz and Kondrich, Andrew and Konstantinidis, Aris and Kosic, Kyle and Krueger, Gretchen and Kuo, Vishal and Lampe, Michael and Lan, Ikai and Lee, Teddy and Leike, Jan and Leung, Jade and Levy, Daniel and Li, Chak Ming and Lim, Rachel and Lin, Molly and Lin, Stephanie and Litwin, Mateusz and Lopez, Theresa and Lowe, Ryan and Lue, Patricia and Makanju, Anna and Malfacini, Kim and Manning, Sam and Markov, Todor and Markovski, Yaniv and Martin, Bianca and Mayer, Katie and Mayne, Andrew and McGrew, Bob and McKinney, Scott Mayer and McLeavey, Christine and McMillan, Paul and McNeil, Jake and Medina, David and Mehta, Aalok and Menick, Jacob and Metz, Luke and Mishchenko, Andrey and Mishkin, Pamela and Monaco, Vinnie and Morikawa, Evan and Mossing, Daniel and Mu, Tong and Murati, Mira and Murk, Oleg and Mély, David and Nair, Ashvin and Nakano, Reiichiro and Nayak, Rajeev and Neelakantan, Arvind and Ngo, Richard and Noh, Hyeonwoo and Ouyang, Long and O'Keefe, Cullen and Pachocki, Jakub and Paino, Alex and Palermo, Joe and Pantuliano, Ashley and Parascandolo, Giambattista and Parish, Joel and Parparita, Emy and Passos, Alex and Pavlov, Mikhail and Peng, Andrew and Perelman, Adam and Peres, Filipe de Avila Belbute and Petrov, Michael and Pinto, Henrique Ponde de Oliveira and Michael and Pokorny and Pokrass, Michelle and Pong, Vitchyr H. and Powell, Tolly and Power, Alethea and Power, Boris and Proehl, Elizabeth and Puri, Raul and Radford, Alec and Rae, Jack and Ramesh, Aditya and Raymond, Cameron and Real, Francis and Rimbach, Kendra and Ross, Carl and Rotsted, Bob and Roussez, Henri and Ryder, Nick and Saltarelli, Mario and Sanders, Ted and Santurkar, Shibani and Sastry, Girish and Schmidt, Heather and Schnurr, David and Schulman, John and Selsam, Daniel and Sheppard, Kyla and Sherbakov, Toki and Shieh, Jessica and Shoker, Sarah and Shyam, Pranav and Sidor, Szymon and Sigler, Eric and Simens, Maddie and Sitkin, Jordan and Slama, Katarina and Sohl, Ian and Sokolowsky, Benjamin and Song, Yang and Staudacher, Natalie and Such, Felipe Petroski and Summers, Natalie and Sutskever, Ilya and Tang, Jie and Tezak, Nikolas and Thompson, Madeleine B. and Tillet, Phil and Tootoonchian, Amin and Tseng, Elizabeth and Tuggle, Preston and Turley, Nick and Tworek, Jerry and Uribe, Juan Felipe Cerón and Vallone, Andrea and Vijayvergiya, Arun and Voss, Chelsea and Wainwright, Carroll and Wang, Justin Jay and Wang, Alvin and Wang, Ben and Ward, Jonathan and Wei, Jason and Weinmann, C. J. and Welihinda, Akila and Welinder, Peter and Weng, Jiayi and Weng, Lilian and Wiethoff, Matt and Willner, Dave and Winter, Clemens and Wolrich, Samuel and Wong, Hannah and Workman, Lauren and Wu, Sherwin and Wu, Jeff and Wu, Michael and Xiao, Kai and Xu, Tao and Yoo, Sarah and Yu, Kevin and Yuan, Qiming and Zaremba, Wojciech and Zellers, Rowan and Zhang, Chong and Zhang, Marvin and Zhao, Shengjia and Zheng, Tianhao and Zhuang, Juntang and Zhuk, William and Zoph, Barret},
	month = mar,
	year = {2024},
	note = {arXiv:2303.08774 [cs]},
}

@misc{mialon_gaia_2023,
	title = {{GAIA}: a benchmark for {General} {AI} {Assistants}},
	shorttitle = {{GAIA}},
	url = {http://arxiv.org/abs/2311.12983},
	doi = {10.48550/arXiv.2311.12983},
	urldate = {2026-01-05},
	publisher = {arXiv},
	author = {Mialon, Grégoire and Fourrier, Clémentine and Swift, Craig and Wolf, Thomas and LeCun, Yann and Scialom, Thomas},
	month = nov,
	year = {2023},
	note = {arXiv:2311.12983 [cs]},
}

@misc{krishna_fact_2025,
	title = {Fact, {Fetch}, and {Reason}: {A} {Unified} {Evaluation} of {Retrieval}-{Augmented} {Generation}},
	shorttitle = {Fact, {Fetch}, and {Reason}},
	url = {http://arxiv.org/abs/2409.12941},
	doi = {10.48550/arXiv.2409.12941},
	urldate = {2026-01-05},
	publisher = {arXiv},
	author = {Krishna, Satyapriya and Krishna, Kalpesh and Mohananey, Anhad and Schwarcz, Steven and Stambler, Adam and Upadhyay, Shyam and Faruqui, Manaal},
	month = jan,
	year = {2025},
	note = {arXiv:2409.12941 [cs]
version: 3},
}

@misc{wu_webwalker_2025,
	title = {{WebWalker}: {Benchmarking} {LLMs} in {Web} {Traversal}},
	shorttitle = {{WebWalker}},
	url = {http://arxiv.org/abs/2501.07572},
	doi = {10.48550/arXiv.2501.07572},
	urldate = {2026-01-05},
	publisher = {arXiv},
	author = {Wu, Jialong and Yin, Wenbiao and Jiang, Yong and Wang, Zhenglin and Xi, Zekun and Fang, Runnan and Zhang, Linhai and He, Yulan and Zhou, Deyu and Xie, Pengjun and Huang, Fei},
	month = aug,
	year = {2025},
	note = {arXiv:2501.07572 [cs]},
}

@inproceedings{vaswani_attention_2017,
	title = {Attention is {All} you {Need}},
	volume = {30},
	url = {https://proceedings.neurips.cc/paper_files/paper/2017/hash/3f5ee243547dee91fbd053c1c4a845aa-Abstract.html},
	urldate = {2026-01-01},
	booktitle = {Advances in {Neural} {Information} {Processing} {Systems}},
	publisher = {Curran Associates, Inc.},
	author = {Vaswani, Ashish and Shazeer, Noam and Parmar, Niki and Uszkoreit, Jakob and Jones, Llion and Gomez, Aidan N and Kaiser, Ł ukasz and Polosukhin, Illia},
	year = {2017},
}

@inproceedings{kim_re-rag_2024,
	address = {Miami, Florida, USA},
	title = {{RE}-{RAG}: {Improving} {Open}-{Domain} {QA} {Performance} and {Interpretability} with {Relevance} {Estimator} in {Retrieval}-{Augmented} {Generation}},
	shorttitle = {{RE}-{RAG}},
	url = {https://aclanthology.org/2024.emnlp-main.1236/},
	doi = {10.18653/v1/2024.emnlp-main.1236},
	urldate = {2026-01-01},
	booktitle = {Proceedings of the 2024 {Conference} on {Empirical} {Methods} in {Natural} {Language} {Processing}},
	publisher = {Association for Computational Linguistics},
	author = {Kim, Kiseung and Lee, Jay-Yoon},
	editor = {Al-Onaizan, Yaser and Bansal, Mohit and Chen, Yun-Nung},
	month = nov,
	year = {2024},
	pages = {22149--22161},
}

@misc{shaoDeepSeekMathPushingLimits2024,
  title = {{{DeepSeekMath}}: {{Pushing}} the {{Limits}} of {{Mathematical Reasoning}} in {{Open Language Models}}},
  shorttitle = {{{DeepSeekMath}}},
  author = {Shao, Zhihong and Wang, Peiyi and Zhu, Qihao and Xu, Runxin and Song, Junxiao and Bi, Xiao and Zhang, Haowei and Zhang, Mingchuan and Li, Y. K. and Wu, Y. and Guo, Daya},
  year = 2024,
  month = apr,
  number = {arXiv:2402.03300},
  eprint = {2402.03300},
  primaryclass = {cs},
  publisher = {arXiv},
  doi = {10.48550/arXiv.2402.03300},
  archiveprefix = {arXiv}
}

@misc{chenReSearchLearningReason2025,
  title = {{{ReSearch}}: {{Learning}} to {{Reason}} with {{Search}} for {{LLMs}} via {{Reinforcement Learning}}},
  shorttitle = {{{ReSearch}}},
  author = {Chen, Mingyang and Sun, Linzhuang and Li, Tianpeng and Sun, Haoze and Zhou, Yijie and Zhu, Chenzheng and Wang, Haofen and Pan, Jeff Z. and Zhang, Wen and Chen, Huajun and Yang, Fan and Zhou, Zenan and Chen, Weipeng},
  year = 2025,
  month = sep,
  number = {arXiv:2503.19470},
  eprint = {2503.19470},
  primaryclass = {cs},
  publisher = {arXiv},
  doi = {10.48550/arXiv.2503.19470},
  archiveprefix = {arXiv}
}

@misc{jinSearchR1TrainingLLMs2025,
  title = {Search-{{R1}}: {{Training LLMs}} to {{Reason}} and {{Leverage Search Engines}} with {{Reinforcement Learning}}},
  shorttitle = {Search-{{R1}}},
  author = {Jin, Bowen and Zeng, Hansi and Yue, Zhenrui and Yoon, Jinsung and Arik, Sercan and Wang, Dong and Zamani, Hamed and Han, Jiawei},
  year = 2025,
  month = aug,
  number = {arXiv:2503.09516},
  eprint = {2503.09516},
  primaryclass = {cs},
  publisher = {arXiv},
  doi = {10.48550/arXiv.2503.09516},
  archiveprefix = {arXiv}
}

@misc{zhouMEM1LearningSynergize2025,
  title = {{{MEM1}}: {{Learning}} to {{Synergize Memory}} and {{Reasoning}} for {{Efficient Long-Horizon Agents}}},
  shorttitle = {{{MEM1}}},
  author = {Zhou, Zijian and Qu, Ao and Wu, Zhaoxuan and Kim, Sunghwan and Prakash, Alok and Rus, Daniela and Zhao, Jinhua and Low, Bryan Kian Hsiang and Liang, Paul Pu},
  year = 2025,
  month = jul,
  number = {arXiv:2506.15841},
  eprint = {2506.15841},
  primaryclass = {cs},
  publisher = {arXiv},
  doi = {10.48550/arXiv.2506.15841},
  archiveprefix = {arXiv}
}

@misc{chhikaraMem0BuildingProductionReady2025a,
  title = {Mem0: {{Building Production-Ready AI Agents}} with {{Scalable Long-Term Memory}}},
  shorttitle = {Mem0},
  author = {Chhikara, Prateek and Khant, Dev and Aryan, Saket and Singh, Taranjeet and Yadav, Deshraj},
  year = 2025,
  month = apr,
  number = {arXiv:2504.19413},
  eprint = {2504.19413},
  primaryclass = {cs},
  publisher = {arXiv},
  doi = {10.48550/arXiv.2504.19413},
  archiveprefix = {arXiv}
}

@misc{packerMemGPTLLMsOperating2024,
  title = {{{MemGPT}}: {{Towards LLMs}} as {{Operating Systems}}},
  shorttitle = {{{MemGPT}}},
  author = {Packer, Charles and Wooders, Sarah and Lin, Kevin and Fang, Vivian and Patil, Shishir G. and Stoica, Ion and Gonzalez, Joseph E.},
  year = 2024,
  month = feb,
  number = {arXiv:2310.08560},
  eprint = {2310.08560},
  primaryclass = {cs},
  publisher = {arXiv},
  doi = {10.48550/arXiv.2310.08560},
  archiveprefix = {arXiv}
}

@misc{xuAMEMAgenticMemory2025a,
  title = {A-{{MEM}}: {{Agentic Memory}} for {{LLM Agents}}},
  shorttitle = {A-{{MEM}}},
  author = {Xu, Wujiang and Mei, Kai and Gao, Hang and Tan, Juntao and Liang, Zujie and Zhang, Yongfeng},
  year = 2025,
  month = jul,
  number = {arXiv:2502.12110},
  eprint = {2502.12110},
  primaryclass = {cs},
  publisher = {arXiv},
  doi = {10.48550/arXiv.2502.12110},
  archiveprefix = {arXiv}
}

@misc{zhengSynapseTrajectoryasExemplarPrompting2024,
  title = {Synapse: {{Trajectory-as-Exemplar Prompting}} with {{Memory}} for {{Computer Control}}},
  shorttitle = {Synapse},
  author = {Zheng, Longtao and Wang, Rundong and Wang, Xinrun and An, Bo},
  year = 2024,
  month = jan,
  number = {arXiv:2306.07863},
  eprint = {2306.07863},
  primaryclass = {cs},
  publisher = {arXiv},
  doi = {10.48550/arXiv.2306.07863},
  archiveprefix = {arXiv}
}

@misc{zhongMemoryBankEnhancingLarge2023,
  title = {{{MemoryBank}}: {{Enhancing Large Language Models}} with {{Long-Term Memory}}},
  shorttitle = {{{MemoryBank}}},
  author = {Zhong, Wanjun and Guo, Lianghong and Gao, Qiqi and Ye, He and Wang, Yanlin},
  year = 2023,
  month = may,
  number = {arXiv:2305.10250},
  eprint = {2305.10250},
  primaryclass = {cs},
  publisher = {arXiv},
  doi = {10.48550/arXiv.2305.10250},
  archiveprefix = {arXiv}
}

@misc{zhang_deep_2025,
	title = {Deep {Research}: {A} {Survey} of {Autonomous} {Research} {Agents}},
	shorttitle = {Deep {Research}},
	url = {http://arxiv.org/abs/2508.12752},
	doi = {10.48550/arXiv.2508.12752},
	urldate = {2025-12-12},
	publisher = {arXiv},
	author = {Zhang, Wenlin and Li, Xiaopeng and Zhang, Yingyi and Jia, Pengyue and Wang, Yichao and Guo, Huifeng and Liu, Yong and Zhao, Xiangyu},
	month = aug,
	year = {2025},
	note = {arXiv:2508.12752 [cs]},
}

@misc{zheng_deepresearcher_2025,
	title = {{DeepResearcher}: {Scaling} {Deep} {Research} via {Reinforcement} {Learning} in {Real}-world {Environments}},
	shorttitle = {{DeepResearcher}},
	url = {http://arxiv.org/abs/2504.03160},
	doi = {10.48550/arXiv.2504.03160},
	urldate = {2025-12-12},
	publisher = {arXiv},
	author = {Zheng, Yuxiang and Fu, Dayuan and Hu, Xiangkun and Cai, Xiaojie and Ye, Lyumanshan and Lu, Pengrui and Liu, Pengfei},
	month = apr,
	year = {2025},
	note = {arXiv:2504.03160 [cs]},
}

@inproceedings{yao_react_2022,
	title = {React: {Synergizing} reasoning and acting in language models},
	shorttitle = {React},
	url = {https://openreview.net/forum?id=WE_vluYUL-X},
	urldate = {2025-12-12},
	booktitle = {The eleventh international conference on learning representations},
	author = {Yao, Shunyu and Zhao, Jeffrey and Yu, Dian and Du, Nan and Shafran, Izhak and Narasimhan, Karthik R. and Cao, Yuan},
	year = {2022},
}

@inproceedings{hong_data_2025,
	title = {Data interpreter: {An} llm agent for data science},
	shorttitle = {Data interpreter},
	url = {https://aclanthology.org/2025.findings-acl.1016/},
	urldate = {2025-12-12},
	booktitle = {Findings of the {Association} for {Computational} {Linguistics}: {ACL} 2025},
	author = {Hong, Sirui and Lin, Yizhang and Liu, Bang and Liu, Bangbang and Wu, Binhao and Zhang, Ceyao and Li, Danyang and Chen, Jiaqi and Zhang, Jiayi and Wang, Jinlin},
	year = {2025},
	pages = {19796--19821},
}

@misc{islam_mapcoder_2024,
	title = {{MapCoder}: {Multi}-{Agent} {Code} {Generation} for {Competitive} {Problem} {Solving}},
	shorttitle = {{MapCoder}},
	url = {http://arxiv.org/abs/2405.11403},
	doi = {10.48550/arXiv.2405.11403},
	urldate = {2025-12-12},
	publisher = {arXiv},
	author = {Islam, Md Ashraful and Ali, Mohammed Eunus and Parvez, Md Rizwan},
	month = may,
	year = {2024},
	note = {arXiv:2405.11403 [cs]},
}

@misc{zhang_codeagent_2024,
	title = {{CodeAgent}: {Enhancing} {Code} {Generation} with {Tool}-{Integrated} {Agent} {Systems} for {Real}-{World} {Repo}-level {Coding} {Challenges}},
	shorttitle = {{CodeAgent}},
	url = {http://arxiv.org/abs/2401.07339},
	doi = {10.48550/arXiv.2401.07339},
	urldate = {2025-12-12},
	publisher = {arXiv},
	author = {Zhang, Kechi and Li, Jia and Li, Ge and Shi, Xianjie and Jin, Zhi},
	month = aug,
	year = {2024},
	note = {arXiv:2401.07339 [cs]},
}

@inproceedings{ho_verilogcoder_2025,
	title = {Verilogcoder: {Autonomous} verilog coding agents with graph-based planning and abstract syntax tree (ast)-based waveform tracing tool},
	volume = {39},
	shorttitle = {Verilogcoder},
	url = {https://ojs.aaai.org/index.php/AAAI/article/view/32007},
	urldate = {2025-12-12},
	booktitle = {Proceedings of the {AAAI} {Conference} on {Artificial} {Intelligence}},
	author = {Ho, Chia-Tung and Ren, Haoxing and Khailany, Brucek},
	year = {2025},
	note = {Issue: 1},
	pages = {300--307},
}

@misc{liu_lost_2023,
	title = {Lost in the {Middle}: {How} {Language} {Models} {Use} {Long} {Contexts}},
	shorttitle = {Lost in the {Middle}},
	url = {http://arxiv.org/abs/2307.03172},
	doi = {10.48550/arXiv.2307.03172},
	urldate = {2025-12-12},
	publisher = {arXiv},
	author = {Liu, Nelson F. and Lin, Kevin and Hewitt, John and Paranjape, Ashwin and Bevilacqua, Michele and Petroni, Fabio and Liang, Percy},
	month = nov,
	year = {2023},
	note = {arXiv:2307.03172 [cs]},
}

@inproceedings{borgeaud_improving_2022,
	title = {Improving {Language} {Models} by {Retrieving} from {Trillions} of {Tokens}},
	shorttitle = {rag},
	url = {https://proceedings.mlr.press/v162/borgeaud22a.html},
	language = {en},
	urldate = {2025-12-28},
	booktitle = {Proceedings of the 39th {International} {Conference} on {Machine} {Learning}},
	publisher = {PMLR},
	author = {Borgeaud, Sebastian and Mensch, Arthur and Hoffmann, Jordan and Cai, Trevor and Rutherford, Eliza and Millican, Katie and Driessche, George Bm Van Den and Lespiau, Jean-Baptiste and Damoc, Bogdan and Clark, Aidan and Casas, Diego De Las and Guy, Aurelia and Menick, Jacob and Ring, Roman and Hennigan, Tom and Huang, Saffron and Maggiore, Loren and Jones, Chris and Cassirer, Albin and Brock, Andy and Paganini, Michela and Irving, Geoffrey and Vinyals, Oriol and Osindero, Simon and Simonyan, Karen and Rae, Jack and Elsen, Erich and Sifre, Laurent},
	month = jun,
	year = {2022},
	note = {ISSN: 2640-3498},
	pages = {2206--2240},
}

@misc{gao_retrieval-augmented_2024,
	title = {Retrieval-{Augmented} {Generation} for {Large} {Language} {Models}: {A} {Survey}},
	shorttitle = {rag},
	url = {http://arxiv.org/abs/2312.10997},
	doi = {10.48550/arXiv.2312.10997},
	urldate = {2025-12-28},
	publisher = {arXiv},
	author = {Gao, Yunfan and Xiong, Yun and Gao, Xinyu and Jia, Kangxiang and Pan, Jinliu and Bi, Yuxi and Dai, Yi and Sun, Jiawei and Wang, Meng and Wang, Haofen},
	month = mar,
	year = {2024},
	note = {arXiv:2312.10997 [cs]},
}

@inproceedings{lewis_retrieval-augmented_2020,
	title = {Retrieval-{Augmented} {Generation} for {Knowledge}-{Intensive} {NLP} {Tasks}},
	volume = {33},
	shorttitle = {rag},
	url = {https://proceedings.neurips.cc/paper/2020/hash/6b493230205f780e1bc26945df7481e5-Abstract.html},
	urldate = {2025-12-28},
	booktitle = {Advances in {Neural} {Information} {Processing} {Systems}},
	publisher = {Curran Associates, Inc.},
	author = {Lewis, Patrick and Perez, Ethan and Piktus, Aleksandra and Petroni, Fabio and Karpukhin, Vladimir and Goyal, Naman and Küttler, Heinrich and Lewis, Mike and Yih, Wen-tau and Rocktäschel, Tim and Riedel, Sebastian and Kiela, Douwe},
	year = {2020},
	pages = {9459--9474},
}

\appendix

\section{Appendix}

\subsection{Single-Objective Tasks}
To assess the effectiveness of our approach on single-objective settings, we conduct experiments on seven question answering benchmarks: 2WikiMultiHopQA, HotpotQA, Bamboogle, Musique, Natural Questions (NQ), TriviaQA, and PopQA \citep{ho_constructing_2020, yang_hotpotqa_2018, press_measuring_2023, trivedi_musique_2022, kwiatkowski_natural_2019, joshi_triviaqa_2017, mallen_when_2023}. These datasets span diverse domains and are widely used in prior agent-oriented research. For datasets with more than 1k samples, we randomly sample 1k samples for evaluation.

\textbf{Datasets.}
The following are details of these datasets:
\begin{itemize}
  \item \textbf{Natural Questions (NQ)}: a QA dataset whose questions are derived from real anonymized and aggregated queries issued to the Google Search engine.
  \item \textbf{TriviaQA}: a large-scale dataset with compositional questions that often require non-trivial reasoning.
  \item \textbf{PopQA}: 14K questions focusing on long-tail factual knowledge.
  \item \textbf{Bamboogle}: a manually constructed multi-hop QA benchmark in which questions are designed to be difficult to answer with a single search engine call.
  \item \textbf{Musique}: a 25K-question multi-hop QA dataset requiring evidence composition across multiple facts.
  \item \textbf{HotpotQA}: a Wikipedia-based multi-hop dataset where answering requires retrieving and reasoning over multiple supporting documents.
  \item \textbf{2WikiMultiHopQA}: a multi-hop QA dataset combining structured and unstructured evidence, explicitly constructed to necessitate multi-hop reasoning.
\end{itemize}

\textbf{Results.}
As shown in Table~\ref{tab_multi-hop} and Table~\ref{tab_single-hop}, our method achieves strong performance across all benchmarks. On several datasets (e.g., TriviaQA), it surpasses all baselines and reaches SOTA performance, while substantially reducing token consumption. These results suggest that our method can significantly improve performance on long-horizon tasks while maintaining competitiveness on short-horizon tasks, matching or even exceeding agent models trained specifically for single-objective settings.

\begin{table*}[htbp]
  \centering
  \caption{The token consumption for multi-objective tasks of baselines.\textbf{Text with bold} means SOTA.}
  \resizebox{\textwidth}{!}{%
    \begin{tabular}{l|cccccccccc|cc}
    \toprule
    \toprule
    \rowcolor[rgb]{ .949,  .949,  .949} \multicolumn{13}{c}{\textbf{Local Wiki Search}} \\
    \midrule
    \midrule
    \multirow{2}[2]{*}{Model} & \multicolumn{2}{c}{\textbf{2-objective}} & \multicolumn{2}{c}{\textbf{4-objective}} & \multicolumn{2}{c}{\textbf{6-objective}} & \multicolumn{2}{c}{\textbf{8-objective}} & \multicolumn{2}{c|}{\textbf{10-objective}} & \multicolumn{2}{c}{\textbf{Avg}} \\
          & TT    & PT    & TT    & PT    & TT    & PT    & TT    & PT    & TT    & PT    & TT    & PT \\
    \midrule
    Qwen2.5 (ReAct) & 1.90  & 0.38  & 2.80  & 0.51  & 3.97  & 0.63  & 4.89  & 0.76  & 4.66  & 0.77  & 3.64  & 0.61  \\
    Research & 0.45  & 0.26  & 1.58  & 0.52  & 3.18  & 0.76  & 4.63  & 0.91  & 6.62  & 1.11  & 3.29  & 0.71  \\
    DeepResearcher & 0.96  & 0.31  & 2.60  & 0.61  & 4.13  & 0.79  & 5.48  & 0.93  & 8.26  & 1.19  & 4.29  & 0.77  \\
    A-MEM & 1.14  & 0.34  & 2.16  & 0.38  & 2.61  & 0.38  & 3.49  & 0.41  & 3.69  & 0.40  & 2.62  & 0.38  \\
    MEM1  & 0.50  & 0.16  & 0.88  & 0.18  & 1.31  & 0.20  & 1.81  & 0.22  & 2.40  & 0.24  & 1.38  & 0.20  \\
    GRPO (no mem) & 0.45  & 0.26  & 2.29  & 0.62  & 3.91  & 0.84  & 6.32  & 1.07  & 8.96  & 1.28  & 4.39  & 0.81  \\
    Ours  & \textbf{0.32 } & \textbf{0.14 } & \textbf{0.80 } & \textbf{0.17 } & \textbf{1.22 } & \textbf{0.19 } & \textbf{1.61 } & \textbf{0.20 } & \textbf{1.94 } & \textbf{0.21 } & \textbf{1.18 } & \textbf{0.18 } \\
    \midrule
    \midrule
    \rowcolor[rgb]{ .949,  .949,  .949} \multicolumn{13}{c}{\textbf{Online Web Search}} \\
    \midrule
    \midrule
    Qwen2.5 (ReAct) & 0.24  & 0.13  & 4.53  & 0.36  & 2.63  & 0.34  & 3.21  & 0.40  & 5.08  & 0.48  & 3.14  & 0.34  \\
    Research & 0.27  & 0.14  & 1.04  & 0.29  & 1.94  & 0.41  & 3.13  & 0.52  & 4.48  & 0.62  & 2.17  & 0.40  \\
    MEM1  & 0.31  & 0.10  & 0.57  & 0.12  & 0.89  & 0.14  & 1.30  & 0.16  & 1.74  & 0.19  & 0.96  & 0.14  \\
    Ours  & \textbf{0.19 } & \textbf{0.08 } & \textbf{0.56 } & \textbf{0.11 } & \textbf{0.85 } & \textbf{0.12 } & \textbf{1.17 } & \textbf{0.14 } & \textbf{1.53 } & \textbf{0.15 } & \textbf{0.86 } & \textbf{0.12 } \\
    \bottomrule
    \bottomrule
    \end{tabular}%
    }
  \label{tab_multi-obj_token-num}%
\end{table*}%

\setlength{\tabcolsep}{3pt}
\begin{table*}[htbp]
  \centering
  \caption{The results of multi-hop QA datasets.They are the first part of single-objective QA datasets.}
  \resizebox{\textwidth}{!}{%
    \begin{tabular}{l|cccc|cccc|cccc|cccc|cccc}
    \toprule
    \toprule
    \rowcolor[rgb]{ .949,  .949,  .949} \multicolumn{21}{c}{\textbf{Multi-hop QA}} \\
    \midrule
    \midrule
    \multirow{2}[2]{*}{Model} & \multicolumn{4}{c|}{2WikiMultiHopQA} & \multicolumn{4}{c|}{Bamboogle} & \multicolumn{4}{c|}{HotpotQA} & \multicolumn{4}{c|}{Musique}  & \multicolumn{4}{c}{Avg} \\
          & F1    & EM    & TT    & PT    & F1    & EM    & TT    & PT    & F1    & EM    & TT    & PT    & F1    & EM    & TT    & PT    & F1    & EM    & TT    & PT \\
    \midrule
    qwen2.5 (ReAct) & 45.21  & 35.90  & 1.12  & 0.29  & 43.70  & 32.80  & 0.95  & 0.24  & 47.12  & 34.30  & 1.07  & 0.26  & 23.86  & 14.40  & 2.06  & 0.37  & 39.97  & 29.35  & 1.30  & 0.29  \\
    ReSearch & 50.07  & 41.90  & 0.55  & 0.28  & 53.61  & 40.80  & 0.41  & 0.24  & 50.26  & 33.90  & 0.40  & 0.23  & 29.63  & 17.80  & 0.58  & 0.29  & 45.89  & 33.60  & 0.48  & 0.26  \\
    DeepResearcher & 51.44  & 43.90  & 1.24  & 0.34  & 48.48  & 37.60  & 1.22  & 0.30  & 51.96  & 38.40  & 1.01  & 0.29  & 26.55  & 17.00  & 2.54  & 0.44  & 44.61  & 34.23  & 1.50  & 0.34  \\
    GRPO (no mem) & 62.35  & 53.80  & 0.93  & 0.32  & 53.18  & 40.80  & 0.79  & 0.28  & 57.29  & 42.30  & 0.62  & 0.27  & 33.78  & 21.50  & 0.86  & 0.33  & 51.65  & 39.60  & 0.80  & 0.30  \\
    Ours  & 59.17  & 50.20  & 0.37  & 0.15  & 52.90  & 36.80  & 0.29  & 0.14  & 57.64  & 42.90  & 0.34  & 0.15  & 33.48  & 22.10  & 0.39  & 0.15  & 50.80  & 38.00  & 0.35  & 0.15  \\
    \bottomrule
    \bottomrule
    \end{tabular}%
    }
  \label{tab_multi-hop}%
\end{table*}%

\begin{table*}[htbp]
  \centering
  \caption{The results of single-hop QA datasets.They are the second part of single-objective QA datasets.}
  \resizebox{\textwidth}{!}{%
    \begin{tabular}{l|cccc|cccc|cccc|cccc}
    \toprule
    \toprule
    \rowcolor[rgb]{ .949,  .949,  .949} \multicolumn{17}{c}{\textbf{Single-hop QA}} \\
    \midrule
    \midrule
    \multirow{2}[2]{*}{Model} & \multicolumn{4}{c|}{NQ}       & \multicolumn{4}{c|}{PopQA}    & \multicolumn{4}{c|}{TriviaQA} & \multicolumn{4}{c}{Avg} \\
          & F1    & EM    & TT    & PT    & F1    & EM    & TT    & PT    & F1    & EM    & TT    & PT    & F1    & EM    & TT    & PT \\
    \midrule
    qwen2.5 (ReAct) & 49.18  & 36.30  & 1.87  & 0.30  & 47.77  & 40.50  & 1.43  & 0.26  & 62.73  & 52.60  & 1.02  & 0.22  & 53.23  & 43.13  & 1.44  & 0.26  \\
    ReSearch & 52.49  & 38.00  & 0.23  & 0.16  & 52.41  & 43.80  & 0.23  & 0.16  & 62.04  & 50.10  & 0.24  & 0.17  & 55.65  & 43.97  & 0.23  & 0.16  \\
    DeepResearcher & 50.15  & 39.50  & 1.19  & 0.25  & 47.90  & 40.80  & 0.60  & 0.22  & 61.89  & 52.30  & 0.95  & 0.23  & 53.32  & 44.20  & 0.91  & 0.23  \\
    GRPO (no mem) & 56.04  & 43.90  & 0.63  & 0.23  & 53.77  & 46.50  & 1.00  & 0.26  & 64.57  & 54.50  & 0.52  & 0.23  & 58.13  & 48.30  & 0.72  & 0.24  \\
    Ours  & 57.46  & 46.10  & 0.22  & 0.14  & 53.53  & 46.70  & 0.25  & 0.14  & 67.83  & 57.10  & 0.25  & 0.14  & 59.61  & 49.97  & 0.24  & 0.14  \\
    \bottomrule
    \bottomrule
    \end{tabular}%
    }
  \label{tab_single-hop}%
\end{table*}%

\begin{table*}[htbp]
  \centering
  \caption{The results of deep research datasets.}
  \resizebox{\textwidth}{!}{%
    \begin{tabular}{l|cccc|cccc|cccc|cccc}
    \toprule
    \toprule
    \rowcolor[rgb]{ .949,  .949,  .949} \multicolumn{17}{c}{\textbf{Deep Research}} \\
    \midrule
    \midrule
    \multirow{2}[2]{*}{Model} & \multicolumn{4}{c|}{GAIA}     & \multicolumn{4}{c|}{Frames}   & \multicolumn{4}{c|}{WebWalker} & \multicolumn{4}{c}{Avg} \\
          & F1    & EM    & TT    & PT    & F1    & EM    & TT    & PT    & F1    & EM    & TT    & PT    & F1    & EM    & TT    & PT \\
    \midrule
    qwen2.5 (ReAct) & 14.28  & 8.74  & 1.54  & 0.22  & 29.37  & 20.00  & 0.61  & 0.15  & 29.93  & 8.50  & 0.42  & 0.14  & 24.52  & 12.41  & 0.86  & 0.17  \\
    ReSearch & 16.00  & 8.74  & 0.41  & 0.17  & 35.04  & 22.40  & 0.37  & 0.16  & 33.02  & 7.29  & 0.25  & 0.13  & 28.02  & 12.81  & 0.34  & 0.15  \\
    DeepResearcher & 22.44  & 16.50  & 4.94  & 0.33  & 37.69  & 24.40  & 0.44  & 0.17  & 32.10  & 9.31  & 1.16  & 0.19  & 30.75  & 16.74  & 2.18  & 0.23  \\
    Ours & 25.07  & 17.48  & 0.62  & 0.12  & 37.56  & 24.00  & 0.33  & 0.10  & 31.85  & 8.91  & 0.29  & 0.09  & 31.49  & 16.79  & 0.41  & 0.10  \\
    \bottomrule
    \bottomrule
    \end{tabular}%
    }
  \label{tab_deep-research}%
\end{table*}%

\subsection{Deep Research Tasks}
We further evaluate our approach on three deep research benchmarks: GAIA \citep{mialon_gaia_2023}, Frames \citep{krishna_fact_2025}, and WebWalkerQA \cite{wu_webwalker_2025}. In contrast to the above datasets, deep research questions are often constructed from real web search results, and thus are fully out-of-domain (OOD) relative to our training setting. This evaluation tests whether our method generalizes to complex OOD tasks. To reduce evaluation cost, for the larger benchmarks (Frames and WebWalkerQA), we randomly sample 250 instances.

\textbf{Datasets.}
\begin{itemize}
  \item \textbf{GAIA}: a collection of 165 tasks spanning three difficulty levels (53 Level-1, 86 Level-2, and 26 Level-3), designed to measure tool use and multi-step reasoning.
  \item \textbf{Frames}: measures multi-perspective reasoning and role-conditioned information synthesis, requiring consistent integration of evidence across different contextual frames.
  \item \textbf{WebWalkerQA}: evaluates complex, multi-turn web interaction, consisting of 680 real-world queries across four domains and over 1{,}373 webpages.
\end{itemize}

\textbf{Results.}
As reported in Table~\ref{tab_deep-research}, in terms of average accuracy, our method achieves performance comparable to DeepResearcher while significantly reducing token usage. Notably, DeepResearcher is trained specifically in real web search environments. Moreover, on longer-horizon tasks with higher token demands (e.g., GAIA), our method delivers relatively strong performance. Overall, these findings indicate that our approach generalizes well to out-of-domain settings while still demonstrating solid capability on long-horizon reasoning tasks.

\begin{table*}[htbp]
  \centering
  \caption{Training prompt text with cumulative summary memory.}
    \begin{tabularx}{\textwidth}{X}
    \toprule
    \toprule
    \rowcolor[rgb]{ .929,  .929,  .929} \textbf{Prompt:}\newline{}\newline{}You will answer multiple complex questions using iterative reasoning, summarization, and web search. At each step, you will see the questions, a cumulative summary of relevant information, the current search query, and search results (except in the first step, where only the questions are provided).\newline{}Your task is to:\newline{}\newline{}1. Update a cumulative, concise summary and perform reasoning within <mem>\textbackslash{}n...\textbackslash{}n</mem>\textbackslash{}n\newline{}<think>\textbackslash{}n...\textbackslash{}n</think>, respectively. <mem>\textbackslash{}n...\textbackslash{}n</mem> must include all essential information from previous <think> and <tool\_response> tags.\newline{}\newline{}2. Then choose one of the following actions:\newline{}   - If any question remains unanswered, issue a single query for one question inside <search>\textbackslash{}n...\textbackslash{}n</search>. \newline{}   - If all questions are answered, provide the final answers separated by semicolons within <answer>\textbackslash{}n answer1; answer2; ... \textbackslash{}n</answer>. The answers must be concise, usually short phrases or words, and avoid any explanations.\newline{}\newline{}Important:\newline{}- The summaries only contain concise and essential information related to the user questions from previous trajectories, but do not include any your own internal knowledge or reasoning content.\newline{}- Must strictly follow one of these two structures: <mem>\textbackslash{}n...\textbackslash{}n</mem>\textbackslash{}n<think>\textbackslash{}n...\textbackslash{}n</think>\textbackslash{}n\newline{}<search>\textbackslash{}n...\textbackslash{}n</search> or <mem>\textbackslash{}n...\textbackslash{}n</mem>\textbackslash{}n<think>\textbackslash{}n...\textbackslash{}n</think>\textbackslash{}n<answer>\textbackslash{}n...\textbackslash{}n</answer>.\newline{}- Do not search multiple queries or questions simultaneously. Don't give up searching for information until you find clear information that provides the answer. \\
    \bottomrule
    \bottomrule
    \end{tabularx}%
  \label{tab_prompt}%
\end{table*}%

\begin{table*}[htbp]
  \centering
  \caption{Case study of our method.}
    \begin{tabularx}{\linewidth}{X}
    \toprule
    \toprule
    \rowcolor[rgb]{ .867,  .922,  .969}
    \textbf{Query}\\
    As of Aug 3, 2024, the artist who released the album "Father of Asahd" went to the same high school as an Olympic diver. How many Olympic teams did this diver participate on?\\
    \midrule
    \rowcolor[rgb]{ .988,  .894,  .839}
    \textbf{Answer}\\
    2\\
    \midrule
    \rowcolor[rgb]{ .929,  .929,  .929}
    \textbf{Step 1}\\
    \textcolor[rgb]{ .267,  .447,  .769}{Assistant:}\\
    <mem>Need to identify the artist, the diver, and the number of Olympic teams.</mem>\\
    <think>To answer the question, I first need to find the artist of "Father of Asahd" ...</think>\\
    <search>Who released the album "Father of Asahd"</search>\\
    \textcolor[rgb]{ .267,  .447,  .769}{User:}\\
    <information>
    [Doc 1-5] ... The album "Father of Asahd" is the 11th studio album by American DJ and producer DJ Khaled, released in May 2019 ...
    </information> \\
    \midrule
    \rowcolor[rgb]{ .929,  .929,  .929}
    \textbf{Step 2}\\
    \textcolor[rgb]{ .267,  .447,  .769}{Assistant:}\\
    <mem>DJ Khaled released "Father of Asahd". He attended the same high school as an Olympic diver.</mem>\\
    <think>Next, I need to confirm DJ Khaled's high school to find the related Olympic diver ...</think>\\
    <search>DJ Khaled high school</search>\\
    \textcolor[rgb]{ .267,  .447,  .769}{User:}\\
    <information>
    [Doc 1-5] ... DJ Khaled graduated from Dr. Phillips High School ...
    </information> \\
    \midrule
    \rowcolor[rgb]{ .929,  .929,  .929}
    \textbf{Step 3}\\
    \textcolor[rgb]{ .267,  .447,  .769}{Assistant:}\\
    <mem>DJ Khaled graduated from Dr. Phillips High School. Need to find the Olympic diver from this school.</mem>\\
    <think>I will search for Olympic divers who attended Dr. Phillips High School ...</think>\\
    <search>Olympic diver who attended Dr. Phillips High School</search>\\
    \textcolor[rgb]{ .267,  .447,  .769}{User:}\\
    <information>
    [Doc 1-5] ... Mark Ruiz, a graduate of Dr. Phillips High School, was a member of the US Olympic diving team at the 2000 and 2004 Summer Olympics ...
    </information> \\
    \midrule
    \rowcolor[rgb]{ .929,  .929,  .929}
    \textbf{Step 4}\\
    \textcolor[rgb]{ .267,  .447,  .769}{Assistant:}\\
    <mem>Olympic diver Mark Ruiz attended Dr. Phillips High School with DJ Khaled. He joined the 2000 and 2004 US Olympic diving teams.</mem>\\
    <think>Mark Ruiz participated in two Olympic Games (2000 and 2004), so he was on 2 Olympic teams ...</think>\\
    <answer>2</answer>
    \\
    \bottomrule
    \bottomrule
    \end{tabularx}%
  \label{tab_case-study_ours}%
\end{table*}%

\begin{table*}[htbp]
  \centering
  \caption{Case study of ReAct. It was failed because of the Hallucination in Step 3. Greg Louganis is also a diver, but he is not mentioned anywhere in the context.}
    \begin{tabularx}{\linewidth}{X}
    \toprule
    \toprule
    \rowcolor[rgb]{ .867,  .922,  .969}
    \textbf{Query}\\
    As of Aug 3, 2024, the artist who released the album "Father of Asahd" went to the same high school as an Olympic diver. How many Olympic teams did this diver participate on?\\
    \midrule
    \rowcolor[rgb]{ .988,  .894,  .839}
    \textbf{Answer}\\
    2\\
    \midrule
    \rowcolor[rgb]{ .929,  .929,  .929}
    \textbf{Step 1}\\
    \textcolor[rgb]{ .267,  .447,  .769}{Assistant:}\\
    
    <think>First, identify the artist who released the album "Father of Asahd" ...</think>
    \\
    <search>
    Who is the artist that released the album "Father of Asahd"?
    </search>\\
    \textcolor[rgb]{ .267,  .447,  .769}{User:}\\
    <information>
    [Doc 1-5] ... Father of Asahd is the eleventh studio album by American DJ and producer DJ Khaled, released on May 17, 2019 ...
    </information> \\
    \midrule
    \rowcolor[rgb]{ .929,  .929,  .929}
    \textbf{Step 2}\\
    \textcolor[rgb]{ .267,  .447,  .769}{Assistant:}\\
    
    <think>The artist who released the album "Father of Asahd" is DJ Khaled ...</think>
    \\
    <search>
    Which high school did DJ Khaled go to?
    </search>\\
    \textcolor[rgb]{ .267,  .447,  .769}{User:}\\
    <information>
    [Doc 1-5] ... DJ Khaled graduated from Dr. Phillips High School and also visited New Rochelle High School for a graduation event ...
    </information> \\
    \midrule
    \rowcolor[rgb]{ .929,  .929,  .929}
    \textbf{Step 3}\\
    \textcolor[rgb]{ .267,  .447,  .769}{Assistant:}\\
    
    <think>DJ Khaled graduated from Dr. Phillips High School ...</think>
    \\
    <search>
    Which high school did Greg Louganis go to?
    </search>\\
    \textcolor[rgb]{ .267,  .447,  .769}{User:}\\
    <information>
    [Doc 1-5] ... Greg Louganis attended Santa Ana High School, Valhalla High School, and Mission Viejo High School in California ...
    </information> \\
    \midrule
    \rowcolor[rgb]{ .929,  .929,  .929}
    \textbf{Step 4}\\
    \textcolor[rgb]{ .267,  .447,  .769}{Assistant:}\\
    
    <think>Greg Louganis attended Santa Ana High School, Valhalla High School, and Mission Viejo High School. He went to Valhalla High School ...</think>
    \\
    <search>
    How many Olympic teams did Greg Louganis participate in?
    </search>\\
    \textcolor[rgb]{ .267,  .447,  .769}{User:}\\
    <information>
    ... [Doc 1-5] Greg Louganis is a four-time Olympian, five-time Olympic medalist and four-time Olympic gold medalist diver ...
    </information> \\
    \midrule
    \rowcolor[rgb]{ .929,  .929,  .929}
    \textbf{Step 5}\\
    \textcolor[rgb]{ .267,  .447,  .769}{Assistant:}\\
    <think>
    Greg Louganis was a four-time Olympian ...
    </think>
    \\
    <answer>
    4
    </answer>
    \\
    \bottomrule
    \bottomrule
    \end{tabularx}%
  \label{tab_case-study_ours_2}%
\end{table*}%

\end{document}